\documentclass[10pt,twocolumn,letterpaper]{article}

\usepackage[pagenumbers]{iccv} 

\usepackage{amsmath}
\usepackage{amssymb}
\usepackage{mathtools}
\usepackage{amsthm}
\usepackage{wrapfig,lipsum,booktabs}
\usepackage{adjustbox} 
\theoremstyle{plain}
\newtheorem{theorem}{Theorem}[section]
\newtheorem{proposition}[theorem]{Proposition}

\theoremstyle{definition}

\theoremstyle{remark}

\usepackage{multirow}
\usepackage{amssymb}
\usepackage{pifont}
\newcommand{\cmark}{\ding{52}}%
\newcommand{\xmark}{\ding{55}}%
\usepackage{enumitem}
\let\oldding\ding
\renewcommand{\ding}[2][1]{\scalebox{#1}{\oldding{#2}}}

\usepackage{graphicx}
\usepackage{subcaption}

\definecolor{iccvblue}{rgb}{0.21,0.49,0.74}
\usepackage[pagebackref,breaklinks,colorlinks,allcolors=iccvblue]{hyperref}

\def\confName{ICCV}
\def\confYear{2025}

\title{QuEST: Low-bit Diffusion Model Quantization via Efficient Selective Finetuning}

\author{%
  Haoxuan Wang$^{1}$\qquad
  Yuzhang Shang$^{2}$\qquad
  Zhihang Yuan$^{4}$\qquad
  Junyi Wu$^{1}$\qquad \\
  Junchi Yan$^{3}$\qquad
  Yan Yan$^{1}$\footnotemark[2]\\[4pt]
  \normalsize
  $^{1}$University of Illinois Chicago\quad
  $^{2}$University of Central Florida\quad
  $^{3}$Shanghai Jiao Tong University\quad
  $^{4}$Houmo AI\\[2pt]
}

\begin{document}
\maketitle
\begin{abstract}
The practical deployment of diffusion models is still hindered by the high memory and computational overhead. Although quantization paves a way for model compression and acceleration, existing methods face challenges in achieving low-bit quantization efficiently. In this paper, we identify imbalanced activation distributions as a primary source of quantization difficulty, and propose to adjust these distributions through weight finetuning to be more quantization-friendly. We provide both theoretical and empirical evidence supporting finetuning as a practical and reliable solution. Building on this approach, we further distinguish two critical types of quantized layers: those responsible for retaining essential temporal information and those particularly sensitive to bit-width reduction. By selectively finetuning these layers under both local and global supervision, we mitigate performance degradation while enhancing quantization efficiency.
Our method demonstrates its efficacy across three high-resolution image generation tasks, obtaining state-of-the-art performance across multiple bit-width settings.
Code is available at \url{https://github.com/hatchetProject/QuEST}.
\end{abstract}    
\renewcommand{\thefootnote}{\fnsymbol{footnote}}
\footnotetext[2]{Corresponding author}
\section{Introduction}
\label{Introduction}
Diffusion models \cite{guided_diffusion,ddpm,ldm,diffusion_survey2} have recently achieved remarkable success in image generation.
However, this success comes at the cost of two major obstacles that limit their efficiency \cite{diffusion_survey}.
The first obstacle is the 
denoising process which requires hundreds to thousands of inference time steps, 
slowing down the generation speed drastically.
The other is the increasing model size, driven by demands for better image fidelity and higher image resolutions. Both factors contribute to considerable latency and increased computational requirements, impeding the application of diffusion models to real-world settings where both time and computational power are carefully restricted.

Neural network quantization offers a feasible solution for accelerating inference speed and reducing memory consumption simultaneously \cite{quantization_survey}, making it a natural solution for deploying diffusion models efficiently. It aims to compress high-bit model parameters into low-bit approximations with negligible performance degradation. For example, 4-bit weight and 4-bit activation quantization can achieve up to 8$\times$ inference time speedup and memory reduction theoretically \cite{quantization_survey2}.
Hence, low-bit quantization of diffusion models emerges as a viable approach for efficiency enhancement.
Unfortunately, existing diffusion model quantization methods that perform well at higher bit-widths face significant limitations in low-bit settings: some only adjust the quantization parameters and fail under low-bit conditions \cite{qdiffusion,ptq4dm,ptqd}, while others succeed but require substantial computational resources comparable to training a diffusion model from scratch \cite{qdm,bitfusion}.
In this work, we aim for efficient low-bit quantization, thereby circumventing the latter choice of resource-intensive training. 


\begin{table}[t!]
\centering
\small
\begin{sc}
\adjustbox{width=1.0\linewidth}{
\begin{tabular}{ccccc}
\toprule
Method & Data & Time \& Memory & Low-bit & Fully \\
       & Free & Efficient   & Compatible & Quantized\\
\midrule
PTQ~\cite{qdiffusion}    & \cmark & \cmark & \xmark & \cmark\\
QAT~\cite{qdm}    & \xmark & \xmark & \cmark & \cmark \\
EfficientDM~\cite{efficientdm} & \cmark & \cmark & \cmark & \xmark \\
\textbf{Ours} & \cmark & \cmark & \cmark & \cmark \\
\bottomrule
\end{tabular}
}
\end{sc}
\vspace{-8pt}
\caption{\textbf{Comparison with different frameworks.} Our method is both efficient and effective for low-bit diffusion model quantization, also achieving a reduced overall bit-width.}
\label{tab:difference}
\vspace{-20pt}
\end{table}

\begin{figure*}[t]
\begin{center}
\centerline{\includegraphics[trim=440 160 460 120,clip,width=0.98\linewidth]{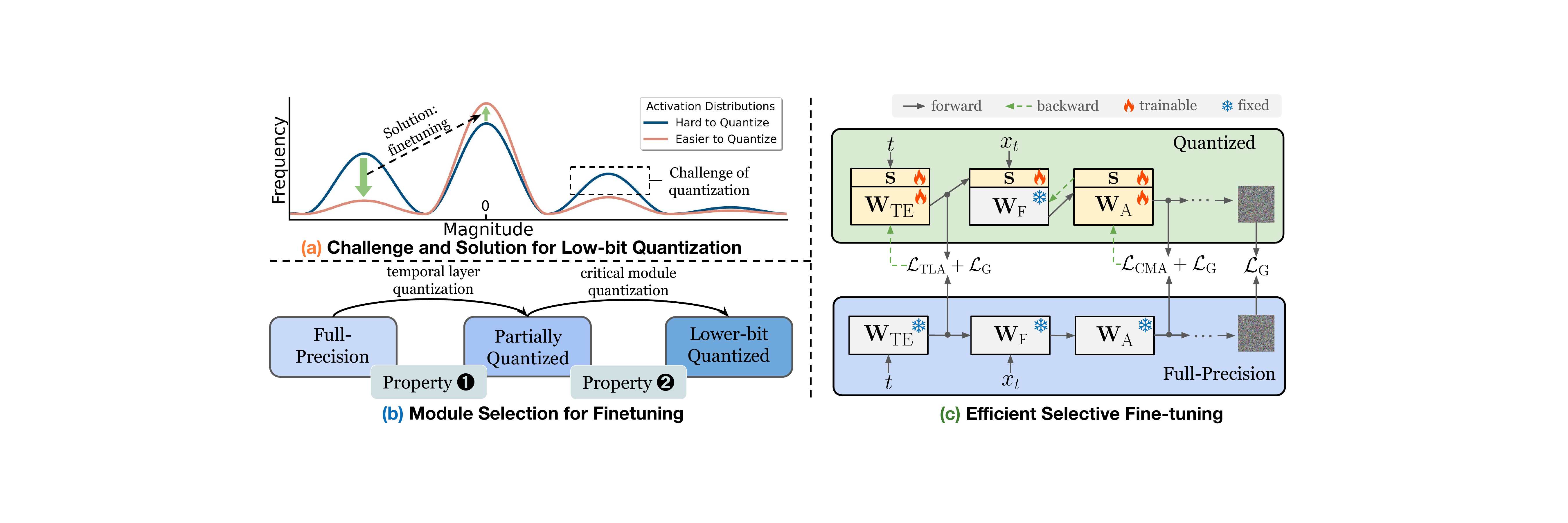}}
\vspace{-10pt}
\caption{Overview of our observations and method. 
\textbf{\color{Orange}{(a)}} Illustration of the challenge we identified in low-bit diffusion model quantization and a potential solution. We propose to ease the quantization difficulty by refining the activation distribution to make it more quantization-friendly.
\textbf{\color{NavyBlue}{(b)}} Overview of property \raisebox{-1,1pt}{\ding[1.1]{182\relax}} and property \raisebox{-1.1pt}{\ding[1.1]{183\relax}}, whose importance are identified based on their impact on the model's generation performance, making them suitable module candidates for efficient finetuning. 
\textbf{\color{OliveGreen}{(c)}} Framework of the proposed method. $\textbf{W}_\text{TE}, \textbf{W}_\text{A}, \textbf{W}_\text{F}$ are the weights of the time embeddings layers, attention-related layers, and other frozen layers, respectively. $\textbf{s}$ is the quantization parameter. To train with efficiency, we adopt a selective and progressive finetuning strategy, incorporating temporal layer alignment (TLA) and critical module alignment (CMA). A global loss is also used for network-level guidance, improving the generated image quality.}
\label{fig:method_flow}
\end{center}
\vspace{-0.4in}
\end{figure*}

We first reveal the current challenge within diffusion models that impede the effectiveness of current efficient low-bit quantization methods \cite{qdiffusion,tdqd,tfmq}. As illustrated in \cref{fig:method_flow}(a): activation distributions tend to be imbalanced, with most values clustering near zero, while essential high-magnitude values are sparse and inconsistently distributed. Existing quantization methods \cite{ptq4dm,qdiffusion,brecq} either approximate large and sparse values, inadequately estimating numerous small values, or focus on small values while overlooking the large ones, thereby impeding the reduction of quantization error. To overcome this challenge, we propose to adjust the activation distributions via weight finetuning, where its feasibility is justified both theoretically and empirically. Nevertheless, finetuning the entire diffusion model is a highly computationally-expensive and time-consuming process, requiring over 80GB memory and numerous hours \cite{qlora,qdm}. Thus, developing an efficient finetuning strategy tailored for diffusion model quantization is important.

To facilitate efficient quantization, we further identify two key properties of quantized diffusion models that unlock new opportunities:
\raisebox{-0.6pt}{\ding[1.1]{182\relax}} diffusion models exhibit varying functions at distinct time steps \cite{perceptiondm}, 
therefore preserving accurate temporal information is important during quantization;
and \raisebox{-0.6pt}{\ding[1.1]{183\relax}} diffusion models possess complex network architectures, incorporating various types of modules. 
Whereas previous works consider each module as equally important and apply quantization uniformly, we reveal that certain modules are particularly sensitive to perturbations from quantization, while others are more resilient. 

Based on the above findings, we propose a novel quantization approach for diffusion models, termed QuEST (\underline{Qu}antization via \underline{E}fficient \underline{S}elective Fine\underline{T}uning). Confronting the revealed quantization challenge, we first theoretically justify that weight finetuning can enhance model robustness toward large activation perturbations in low-bit settings, thereby reducing quantization error. In contrast, previous methods have struggled to properly balance clipping error and rounding error. Then we empirically demonstrate that by finetuning the model weights, the activation distributions are modified to be more amenable to quantization. As shown in \cref{fig:method_flow}(a), the activation distribution is adjusted by reducing the amount of large, sparse values and enhancing the compactness of the value distribution.

Following the idea of weight finetuning, we compare the effects of quantizing different modules of diffusion models (\cref{fig:method_flow}(b)) and identify two types of layers as primary culprits to performance degradation: time embedding layers exhibiting property \raisebox{-1.1pt}{\ding[1.1]{182\relax}} and attention-related layers associated with property \raisebox{-1.1pt}{\ding[1.1]{183\relax}}. Consequently, we selectively and progressively finetune the small subsets of identified layers in conjunction with all activation quantization parameters, as illustrated in \cref{fig:method_flow}(c). The learning objective is crafted to align the quantized model with its full-precision counterpart at both local and global levels.
Involving less than 7\% of the total parameters, QuEST not only substantially enhances low-bit quantized model performance, but is also notably time-efficient and can be conducted in a data-free manner. Our contributions are summarized as follows:
\begin{itemize}[leftmargin=10pt,topsep=0pt,itemsep=1pt,partopsep=1pt,parsep=1pt]
    \item We identify the current challenge in low-bit diffusion model quantization that hinders effective low-bit quantization, and propose to adjust the activation distributions via weight finetuning for easier quantization.
    Both theoretical and empirical discussions are provided.
    \item We uncover and validate two properties in quantized diffusion models as the main factors for degraded performance. Motivated by the identified properties, we introduce QuEST, a parameter-efficient finetuning strategy that trains the diffusion model selectively and progressively, achieving low-bit quantization capability with time and memory efficiency.
    \item Experiments on three high-resolution image generation tasks over four models demonstrate the superiority of our method, achieving state-of-the-art performance under various bit-width settings. 
\end{itemize}

\section{Related Works}
\subsection{Diffusion Model Inference}
Diffusion models \cite{ddpm,ldm,diff_img_trans} generate samples via an iterative denoising process. During inference, the initial input is sampled from a Gaussian distribution: $x_{T} \sim \mathcal{N}(\mathbf{0}, \mathbf{I})$, and the final output $x_{0}$ is obtained through a denoising process: 
\begin{align}
\label{eq:1}
    p_{\theta}(x_{t-1}|x_{t}) = \mathcal{N}(x_{t-1}; \tilde{\boldsymbol{\mu}}_{\theta,t}(x_{t}), \tilde{\beta_{t}}\mathbf{I}),
\end{align}
where $\tilde{\boldsymbol{\mu}}_{\theta,t}$ and $\tilde{\beta_{t}}$ are calculated from the model's output. This denoising process in a typical diffusion model requires tens to thousands of iterations, making efficient inference extremely challenging. 
Practically, diffusion models typically adopt a UNet architecture \cite{unet}, incorporating an encoder and a decoder. Usually, encoders and decoders are lightweight and computationally inexpensive, so our focus is on quantizing the UNet structures in latent diffusion models, in alignment with the other works.

\subsection{Diffusion Model Quantization}
Model quantization is a dominant technique for optimizing the inference memory and speed of deep learning models by reducing the precision of the tensors used in computation. The researches for diffusion model quantization fall into three categories: Quantization-Aware Training (QAT) \cite{qat,qdm,qvit}, Post-Training Quantization (PTQ) \cite{brecq,pdquant,qdrop,ptq4dm,qdiffusion,ptq4dit}, and Parameter-Efficient Fine-Tuning methods \cite{efficientdm}. QAT methods \cite{qdm} train all parameters from scratch, being effective for low-bit quantization but are extremely resource-intensive. PTQ methods \cite{adp-dm,ptqd,tfmq,eqs,tdqd} calculate the quantization parameters based on a small calibration set, offering better efficiency. However, PTQ methods often rely on complex designs and fail at lower bit-widths. To achieve low-bit compatibility with high efficiency, Parameter-Efficient Fine-Tuning methods were proposed. The representative work EfficientDM \cite{efficientdm} trains a low-rank adapter (LoRA) \cite{lora} for each layer to reduce training costs, and successfully scales to W4A4. 

Our proposed method also adopts a parameter-efficient finetuning strategy, and differs from EfficientDM in the following aspects: Firstly, EfficientDM introduces extra weight parameters, requiring substantial training iterations on the LoRA weights. Our method instead does not include additional parameters. Secondly, EfficientDM does not quantize the matrix multiplications in the attention mechanism, as well as certain linear layers. Our method quantizes all layers, and is more time-efficient. \cref{tab:difference} summarizes the differences between our method with the other works. 

\section{Methodology}
\subsection{Preliminaries}
The quantization process for a single value $x$ in a vector can be formulated as:
\begin{align}
\label{eq:quantize}
  \hat{x} &= \text{clamp}(\text{round}(\frac{x}{s}) + Z; q_{min}, q_{max}),
\end{align}
where $\hat{x}$ is the quantized integer result, $\text{round}(\cdot)$ represents rounding algorithms such as the round-to-nearest operator \cite{brecq} and AdaRound \cite{adaround}, $s$ is referred to as the scaling factor and $Z$ is the zero-point. $\text{clamp}$ is the function that clamps values into the range of $[q_{min}, q_{max}]$, which is determined by the bit-width. Reversely, transforming the quantized values back into the full-precision form is:
\begin{align}
\label{eq:dequantize}
    \tilde{x} = (\hat{x} - Z) * s.
\end{align}
This is denoted as the dequantization process. The quantization and dequantization processes are performed on both model weights and layer outputs (also termed as 'activation'). \cref{eq:quantize} indicates that the quantization error is composed of two factors: the clipping error produced by range clamping and the rounding error caused by the rounding function, where they exhibit a trade-off relationship \cite{brecq}. While previous approaches strive for an optimal balance between the two errors, they neglect the intrinsic characteristics in quantized diffusion models. In the following sections, we first examine the current challenges in diffusion model quantization and outline our finetuning motivation, providing theoretical justification. We then identify two properties that enable efficient finetuning, forming the basis of our proposed method.

\begin{figure}[t]
\begin{minipage}[H]{1\linewidth}
\centering
\includegraphics[trim=55 75 55 60,clip,width=0.98\columnwidth]{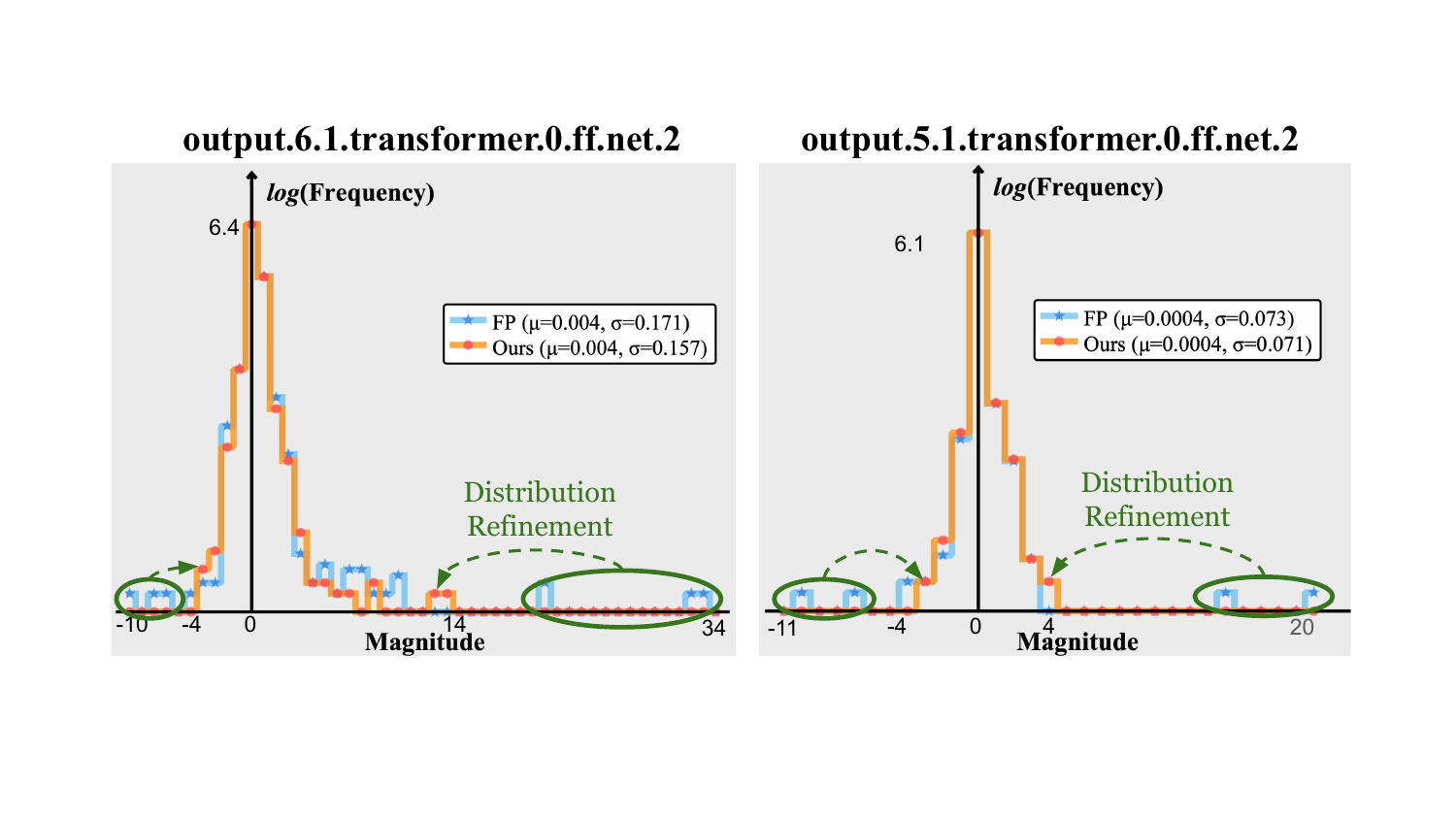}
\vspace{-8pt}
\caption{Illustration of the imbalanced activation distributions. In the full-precision model, the majority of values cluster near zero with sporadic large values, presenting challenges for low-bit quantization. Our method refines the activation distributions by eliminating the large and sparse values, enabling easier quantization.}
\label{fig:activation_range}
\vspace{-15pt}
\end{minipage}
\end{figure}

\subsection{The Challenge in Low-bit Quantization}
\label{sec:3.2}
Previous works \cite{qdiffusion,tfmq} primarily address the varying activation distributions across different time steps, facilitating diffusion model quantization at higher bit-widths. However, these methods experience failure in low-bit settings. To investigate the potential reason, we focus on the activation distribution itself and conduct a layer-wise analysis, revealing the following challenge in full-precision diffusion models that impedes effective quantization:

\noindent\textbf{Challenge: Despite the majority of values being close to zero in the activation outputs, there exist numerically large and sparse values holding significant importance.}

\cref{fig:activation_range} provides a detailed analysis, where activation values are clustered into uniformly distributed bins. We find that in some layers, though the majority of values are close to zero, there exist values that are relatively large and diverse (circled in green). Take the bin plot on the left as an example, the original activation values (blue line) range from [-10, 34] but with most values between [-0.6, 1.7]. Visualizations for more layers can be found in Appendix \ref{sec:example_dist}. This phenomenon poses difficulties in minimizing the clipping error and is unfriendly for effective quantization. 

Moreover, these large and sparse values are important for generation performance preservation. We find that when replacing the few tokens with maximum values by random noises, the generated images' quality is critically degraded (as shown in Appendix \ref{sec:large_act}). With these large values being important and the small values appearing frequently, neither of them is negligible and needs to be carefully quantized at the same time. Unfortunately, typical quantization methods fall short of this ability under low-bit settings, where the rounding error often outweighs the clipping error during optimization and results in over-clipped values, generating corrupted images. This inspires us to refine the activation distributions to attain more quantization-friendly distributions, as depicted in \cref{fig:method_flow}(a).

However, the activation distribution cannot be directly manipulated. To address this, we instead finetune the model weights under quantization constraints, producing a \textit{new yet similar} full-precision model whose quantized counterpart maintains performance comparable to the original full-precision model. Our experiments interestingly reveal that the proposed finetuning strategy effectively eliminates the large and sparse values (\cref{fig:activation_range}), reducing quantization difficulty. We detail our approach in \cref{sec:quest} and provide further theoretical analysis in \cref{sec:th}.


\subsection{Quantization via Efficient Selective Finetuning}
\label{sec:quest}
In this section, we introduce QuEST, an efficient finetuning method for diffusion models that can significantly boost low-bit performance with less time and memory usage. We also present the two unique properties in quantized diffusion models, which serve as the foundation for the design of our method. \cref{fig:method_flow}(c) illustrates our approach.

\subsubsection{Data-free Efficient Network-wise Training.} 
\label{sec:framework}
We first present the general training pipeline of our method. To alleviate the need for substantial training data, we construct the calibration set in a data-free manner. By feeding random Gaussian noises $x_{T}$ into the full-precision model and sampling over different time steps, we can obtain the calibration data needed for finetuning the quantized model. In practice, we only have to infer the full-precision model a few times to gather the needed number of calibration samples, totaling 128 or 256 samples per time step.

As depicted in \cref{fig:method_flow}(c), to overcome the quantization challenge efficiently, we update partial model weights ($\mathbf{W}_{\text{TE}}$ and $\mathbf{W}_{\text{A}}$) that only account for a small subset of parameters related to the time step $t$. The remaining weight parameters $\mathbf{W}_{\text{F}}$ are kept frozen during optimization. We also fix the weight quantization parameters during training, reducing the amount of parameters that need to be optimized. For instance, in LDM-4 \cite{ldm}, no more than 7\% of the parameters are adjusted. The choices for the weights to be finetuned will be discussed in the following sections. 

The activation quantization parameters can be viewed as additional model parameters. Therefore, we further propose a network-wise training strategy. Different from quantization methods using layer-wise or block-wise reconstruction \cite{qdiffusion,ptq4dm} that bind quantization parameters with their corresponding layers or blocks, we optimize all activation scaling factors together with the partial weight parameters.
Additionally, while layer/block-wise optimization methods can only reconstruct sequentially, we update the required parameters at once. In this way, we significantly save the time and memory needed for quantization. 

\subsubsection{Temporal Layer Alignment}
The inference process of diffusion models is highly dependent on the temporal information. Specifically, integer time steps are transformed into time embeddings through one or two linear layers, then added to the intermediate model features. Motivated by this observation, we make the following analysis that is consistent with previous works \cite{tfmq,tdqd}:

\noindent\textbf{\textit{Property \raisebox{-0.5pt}{\ding[1.1]{182\relax}}: Although time embeddings depend solely on time steps and are easily obtainable, precise temporal information is crucial for optimal quantization.}} 

\begin{table}[h]
\centering
\small
\vspace{-8pt}
\adjustbox{width=1.0\linewidth}{
\begin{tabular}{ccc|cccc}
\toprule
\begin{tabular}[c]{@{}c@{}} TE Setting\\ (W8A8 Model)\end{tabular} & FID $\downarrow$ & sFID $\downarrow$ & \begin{tabular}[c]{@{}c@{}} TE Setting\\ (W4A8 Model)\end{tabular} & FID $\downarrow$ & sFID $\downarrow$ \\
\midrule
PTQ & 7.58 & 22.07 & PTQ & 8.59 & 22.74 \\ 
FP & 6.77 & 22.03 & FP & 7.55 & \textbf{21.69} \\ \midrule
\textbf{Ours} & \textbf{5.61} & \textbf{21.22} & \textbf{Ours} & \textbf{6.95} & 23.17 \\
\bottomrule
\end{tabular}
}
\vspace{-8pt}
\caption{Ablations on time embedding (TE) settings. Finetuning the TE layers with our method surpasses full-precision embeddings, while the latter outperforms standard quantized ones.}
\label{tab:time_embed}
\vspace{-8pt}
\end{table}

\cref{tab:time_embed} provides an empirical justification, where we quantitatively show the performance drop when quantizing time embeddings to different bit-width. Under W8A8 and W4A8 bit-width settings, solely quantizing the time embeddings can lead to an increase of 0.81 and 1.04 (relatively 15\%) in FID, respectively. We infer the reason is that inaccurate time embeddings can cause mismatched input and model functionality, resulting in possible oscillations in the sequence of noise removal. Previous works either propose to learn dynamic quantization parameters across different time steps through a simple network \cite{tdqd}, or calibrate the time embedding layers and projection layers across all time steps \cite{tfmq}. We instead focus on finetuning the time embedding layers, adjusting fewer modules without introducing additional parameters. The results in \cref{tab:time_embed} also suggest that our method can improve the quantization performance, even surpassing the full-precision baseline.

Concretely, in a single forward process, identical time embeddings are injected into different parts of the model, passed through projection layers, and merged with the latent image representations. This implies that the time information operates independently from the primary network flow. Thus, we refine the time embedding layer $l$'s weight $\mathbf{w}_{l}$ along with its activation quantization parameters $\mathbf{s}_{l}$:
\begin{align}
\label{eq:time_loss}
\mathcal{L}_{\text{TLA}} = \sum_{l \in \mathbb{C}_{\text{TE}}} \mathbb{E}_{t}[ ||O(t;\mathbf{w}_l) - \tilde{O}(t;\mathbf{w}_l, \mathbf{s}_l)||^{2}], 
\end{align}
where $\mathbb{C}_{\text{TE}}$ represents the set of time embedding layers. $O(t;\mathbf{w}_{l})$ is the intermediate activation of the full-precision model representing the ground truth, and $\tilde{O}(t;\mathbf{w}_{l}, \mathbf{s}_{l})$ is the quantized activation. This objective function indicates that the chosen weight parameters are consistently updated across different time steps, so as to ensure robustness to diverse temporal inputs. Different from other methods \cite{efficientdm,tfmq} that obtain different sets of quantization parameters for each time step, we only use \textbf{a single set} for varying time steps, improving time efficiency and memory storage.

\subsubsection{Critical Module Alignment}
While inaccurate time embedding quantization reduces performance under low-bit settings, it does not cause the complete generation failure observed in fully quantized models. Through careful layer-wise empirical study, we make the following observation:

\noindent\textbf{\textit{Property \raisebox{-0.6pt}{\ding[1.1]{183\relax}}: Not all activations respond equally to reduced bit-width, as different activations exhibit varying levels of sensitivity, with certain critical layers being especially sensitive to quantization.}} 

\begin{figure}[h]
\vspace{-8pt}
\centering
\centerline{\includegraphics[trim=100 50 100 30,clip,width=0.98\columnwidth]{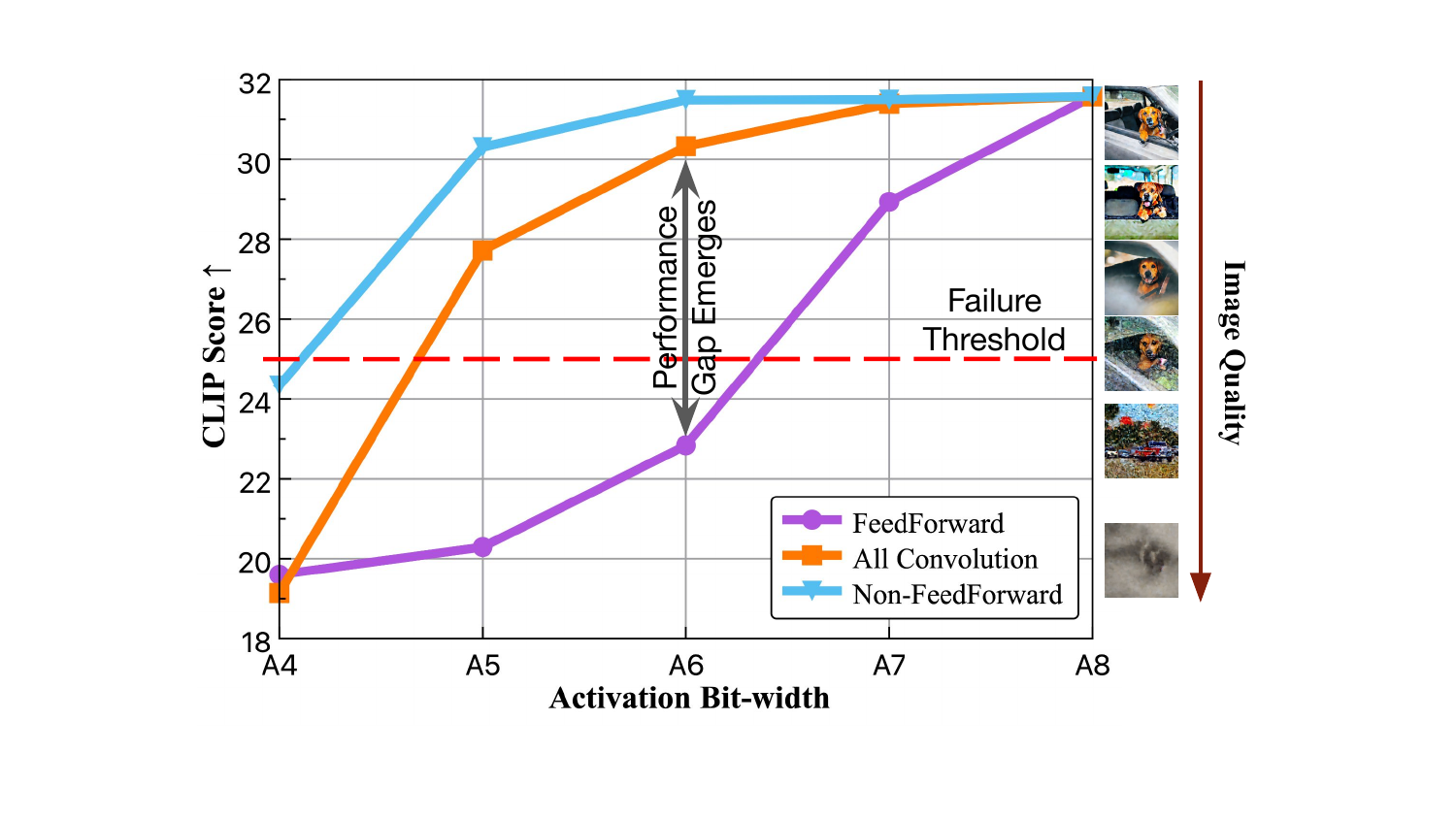}}
\vspace{-8pt}
\caption{Effect of decreasing different activations' bit-width on the model performance. The  generation failure of FeedForward layers emerges at 6 bits, while all other linear layers barely fail at 4 bits and all convolution layers only fail at 4 bits.}
\label{fig:module_importance}
\vspace{-12pt}
\end{figure}

\cref{fig:module_importance} illustrates the sensitivity of different activations to quantization. Specifically, we quantize three different types of activations to lower bits while maintaining the others' bit-width to 8-bit, and observe how the decreasing bit-width affects generation performance. Compared to weights that only fall into linear and convolutional layers, activations are more diverse and complex, making their effective quantization more challenging. Surprisingly, we observe that the FeedForward layer \cite{ffl} activations cause generation failure at as early as 6 bits, whereas the activations of all other linear layers (containing 5 times more layers) barely fail at 4 bits and all convolution layers (containing 3 times more layers) only fail at 4 bits. This indicates that these activations are especially sensitive to low-bit quantization, making them essential to be specially dealt with.

Denote $\mathbb{C}_{\text{A}}$ as the set containing all attention-related \cite{van} layers and given their image calibration inputs $z_{t,l}$, we optimize the corresponding weights and all quantization parameters $\mathbf{s}$, except for the ones already updated:
\begin{align}
\label{eq:qkv_loss}
\mathcal{L}_{\text{CMA}} = \sum_{l \in \mathbb{C}_{A}} \mathbb{E}_{t}[||O(z_{t,l};\mathbf{w}_{l}) - \tilde{O}(\tilde{z}_{t,l};\mathbf{w}_{l}, \mathbf{\hat{s}})||^{2}],  
\end{align}
where $\mathbf{w}_{l}$ are the weight parameters of the $l$th layer, $\tilde{z}_{t,l}$ is the quantized layer input, $\mathbf{\hat{s}}=\mathbf{s} \setminus \mathbf{s}_l, l \in \mathbb{C}_{\text{TE}}$, which represents all the quantization parameters without the ones already finetuned in \cref{eq:time_loss}. Note that we use different inputs to optimize each module, so as to enhance the robustness of the modules to the input perturbations.

\subsubsection{Progressive Alignment with Global Loss} 
As investigated in the previous sections, two crucial types of layers are identified and selected for weight finetuning to enable quantization efficiency: time embedding layers and attention-related layers. We progressively align these components with the full-precision model due to their distinct, non-overlapping functionalities. Since temporal information is independent of the image input and determined early in the model, we first finetune the time embedding layers to provide accurate time step guidance for each subsequent module. Then we optimize the attention-related modules with the refined time embeddings.

However, the above selective finetuning strategy only aligns the local information in the model, but is unaware of the global error reduction of the quantized model and the quantization parameters of the unselected layers. To improve the final generated images' quality, we further aim to minimize the target task loss to provide global supervision:
\begin{align}
\label{eq:global_loss}
\mathcal{L}_{\text{G}} = \mathbb{E}_{t}[||O(x_{t};\mathbf{w}) - \tilde{O}(x_{t};\mathbf{w},\mathbf{s}) ||^{2}],
\end{align}
where $\mathbf{w}$ represents all the model weights, $O(x_{t};\mathbf{w})$ represents the final output of the full-precision counterpart and $\tilde{O}(x_{t};\mathbf{w},\mathbf{s})$ is the final output of the quantized model.

By integrating \cref{eq:time_loss}, \cref{eq:qkv_loss} and \cref{eq:global_loss}, the final objective is formulated as:
\begin{align}
\label{eq:final_loss}
\mkern28mu
\arg\min_{\mathbf{w}_l} (\mathcal{L}_{\text{TLA}} + \mathcal{L}_{\text{CMA}} + 2\mathcal{L}_{\text{G}}), \ \ l \in \mathbb{C}_{\text{TE}} \cup \mathbb{C}_{\text{A}} .
\end{align}

\subsection{Finetuning from a Theoretical Perspective}
\label{sec:th}
The above proposed method is motivated by the intuition that finetuning the model weights can adjust the activation distribution such that the imbalanced activation phenomenon can be alleviated. In this part, we \textit{attempt to explain why finetuning may be a feasible solution}, offering additional insights for readers. However, we note that this is not a theoretical guarantee of the proposed method.

We first review the underlying theory underpinning conventional post-training-quantization methods, which typically employ the reconstruction-based approach. Denote the full-precision diffusion model's activations at time $t$ as $\mathbf{z}_{t}=[z_{1,t}, z_{2,t}, ..., z_{n,t}]$, the final loss as $L(\mathbf{z}_{t}; \mathbf{w})$, where $n$ is the number of layers. $L$ can be any loss function and here we use the mean squared error (MSE). We treat quantization as a type of perturbation and formulate the influence of activation quantization using Taylor expansion, assuming model weight $\mathbf{w}$ is frozen:
\begin{align}
\label{eq:quant_err}
    \mathbb{E}[L(z_{n,t} &+ \Delta; \mathbf{w})] - \mathbb{E}[L(z_{n,t}; \mathbf{w})] \notag \\
    &\approx \Delta^{\mathrm{T}} \overline{\mathbf{g}}^{(z_{n,t})} + \frac{1}{2}\Delta^{\mathrm{T}}\overline{\mathbf{H}}^{(z_{n,t})} \Delta,
\end{align}
where $\Delta$ is the activation perturbation, $\overline{\mathbf{g}}^{(\mathbf{z})}$ is the gradient and $\overline{\mathbf{H}}^{(z_{n,t})}$ is the Hessian matrix. According to \cite{brecq,ptq4vit}, for a well-trained model, $\overline{\mathbf{g}}^{(z_{n,t})} = \nabla_{z_{n,t}}L$ approaches $0$. Thus the above equation can be simplified to:
\begin{align}
\label{eq:brecq}
    \mkern-18mu \frac{1}{2}\Delta^{\mathrm{T}}\overline{\mathbf{H}}^{(z_{n,t})} \Delta = \frac{1}{2}(\tilde{z}_{n,t}-z_{n,t})^{\mathrm{T}}\overline{\mathbf{H}}^{(z_{n,t})}(\tilde{z}_{n,t}-z_{n,t}). \mkern-18mu
\end{align}
However, under low-bit settings, the reasoning from \cref{eq:quant_err} to \cref{eq:brecq} is inaccurate, where the activation perturbation $\Delta$ is too large for a meaningful Taylor expansion. Thus we have the following proposition:

\begin{proposition}
Reconstruction-based post-training quantization methods may lose their theoretical guarantee due to the large value perturbations under low-bit quantization. 
\end{proposition}

Since the inaccuracy arises from the large activation perturbation $\Delta$, we transform $\Delta$ into a smaller perturbation $\epsilon$ and derive the following theorem:

\begin{theorem}
\label{th:weight_finetune}
Given an $n$ layer diffusion model at time $t$ with quantized activations as $\tilde{\mathbf{z}}_{t}=[\tilde{z}_{1,t}, \tilde{z}_{2,t}, ..., \tilde{z}_{n,t}]$ and $\tilde{z}_{n,t}=z_{n,t} + \Delta$, where $z_{n,t}$ is the ground truth and $\Delta$ is the large perturbation caused by low-bit quantization. Denote the target task MSE loss as $L(\mathbf{z}_{t}; \mathbf{w})$, the quantization error can be transformed into:
\begin{align}
    \label{eq:th1}
    & \mathbb{E}[L(z_{n,t} + \Delta; \mathbf{w})] - \mathbb{E}[L(z_{n,t}; \mathbf{w})]  \notag \\
    \approx & 2\epsilon^{\mathrm{T}} \sum_{i=1}^{K} (\tilde{z}_{n-1,t}^{i} \cdot \mathbf{w}_{n} - z_{\text{n,t}}) \notag\\
    & + \frac{1}{2}\sum_{i=1}^{K} (\tilde{z}_{n,t}^{i} - z_{\text{n,t}})^{\mathrm{T}}\overline{\mathbf{H}}^{(z_{n,t}+(i-1)\epsilon)}(\tilde{z}_{n,t}^{i} - z_{\text{n,t}})
\end{align}
where $\mathbf{w}_{n}$ is the weight for layer $n$ and $\tilde{z}_{n,t}^{i}=\tilde{z}_{n-1,t}^{i} \cdot \mathbf{w}_{n}$, $K$ is a large constant and $\Delta=K\epsilon$.
\end{theorem}

\begin{table}[t]
\small
\centering
\vspace{2pt}
\adjustbox{width=1.0\linewidth}{
\begin{tabular}{ccccc}
\toprule
Dataset                                                                    & Method      & \begin{tabular}[c]{@{}c@{}}Bit-width\\ (W/A)\end{tabular}  & \begin{tabular}[c]{@{}c@{}}Size\\ (MB)\end{tabular}  & FID$\downarrow$  \\ \midrule
\multirow{16}{*}{\begin{tabular}[c]{@{}c@{}}LSUN-\\ Bedrooms\\ (LDM-4)\end{tabular}} & \textcolor{gray}{FP}      & \textcolor{gray}{32/32}  &  \textcolor{gray}{1045.6}  & \textcolor{gray}{2.95} \\ \cmidrule{2-5} 
& PTQ4DM      & 8/8   &  279.1   & 4.75 \\
& Q-Diffusion & 8/8   &  279.1   & 4.53 \\
& PTQ-D       & 8/8   &  279.1   & 3.75 \\
& EfficientDM$^*$  & 8/8   & 279.1 & N/A \\
& \textbf{Ours}        & 8/8   &  279.1   & \textbf{3.03} \\ \cmidrule{2-5}
& PTQ4DM      & 4/8   &  148.4  & N/A    \\
& Q-Diffusion & 4/8   &  148.4  & 5.37 \\
& PTQ-D       & 4/8   &  148.4  & 5.94 \\
& EfficientDM$^*$  & 4/8   & 148.4 & 15.15 \\
& \textbf{Ours}        & 4/8   &  148.4  & \bf{3.26} \\ \cmidrule{2-5}
& PTQ4DM      & 4/4   &  148.4   & N/A  \\
& Q-Diffusion & 4/4   &  148.4   & N/A  \\
& PTQ-D       & 4/4   &  148.4   & N/A  \\
& EfficientDM$^*$  & 4/4   & 148.4 & 10.60 \\
& \textbf{Ours}        & 4/4   &  148.4   & \bf{5.64} \\ \cmidrule{1-5}
\multirow{16}{*}{\begin{tabular}[c]{@{}c@{}}LSUN-\\ Churches\\ (LDM-8)\end{tabular}} 
& \textcolor{gray}{FP} & \textcolor{gray}{32/32} & \textcolor{gray}{1125.4} & \textcolor{gray}{4.02} \\ \cmidrule{2-5}
& PTQ4DM*      & 8/8   &  330.6   & 63.93 \\
& Q-Diffusion & 8/8   &  330.6   & 6.94 \\
& PTQ-D*       & 8/8   &  330.6  & 10.76 \\
& EfficientDM$^*$ & 8/8   & 330.6 & N/A \\
& \textbf{Ours}        & 8/8   &  330.6   & \textbf{6.55} \\ \cmidrule{2-5} 
& PTQ4DM*      & 4/8   &  189.9   & N/A    \\
& Q-Diffusion & 4/8   &  189.9  & 7.80 \\
& PTQ-D*       & 4/8   &  189.9   & 7.33 \\
& EfficientDM$^*$  & 4/8   & 189.9  & 9.29 \\
& \textbf{Ours}        & 4/8   &  189.9   & \bf{7.33} \\ \cmidrule{2-5}
& PTQ4DM*      & 4/4   &  189.9   & N/A  \\
& Q-Diffusion & 4/4   &  189.9   & N/A  \\
& PTQ-D*       & 4/4   &  189.9   & N/A  \\
& EfficientDM$^*$   & 4/4   & 189.9 & 14.34 \\
& \textbf{Ours}  & 4/4   &  189.9   & \bf{11.76} \\ \bottomrule
\end{tabular}
}
\vspace{-6pt}
\caption{Quantization performance on LSUN-Bedrooms/Churches 256$\times$256. ``N/A'' denotes generation failure. ``*'' denotes the results obtained by re-implementing the open-source code. More baseline and metric comparisons are included in the Appendix. }
\label{tab:uncond_result}
\vspace{-16pt}
\end{table}

\cref{th:weight_finetune} indicates that, to minimize quantization error, $\mathbf{w}_{n}$ should ideally be fine-tuned so that, for any $i$, the weights fit the corresponding input $\tilde{z}_{n-1,t}^{i}+(i-1)\epsilon$. This adjustment captures variations that the full-precision model may overlook. In other words, \textbf{fine-tuning optimizes model weights for better robustness towards large input activation perturbations, facilitating easier quantization.} Moreover, since the finetuned and quantized model is aligned with the original full-precision model, the potential impact on generation performance can be avoided.
Note that the second term in \cref{eq:th1} can be ignored within an acceptable upper bound, as it is of second order and shares a common zero-loss solution with the first term.

\section{Experiments}
\subsection{Experiment Settings}
To verify the effectiveness of our proposed method, we conduct experiments on three types of generation tasks: Unconditional image generation on LSUN-Bedrooms and LSUN-Churches datasets \cite{lsun}, class-conditional image generation on ImageNet \cite{imagenet}, and text-to-image generation. The model architectures we quantize include LDMs and Stable Diffusion \cite{ldm}, and use "WnAm" to represent the quantization setting: n-bit weight quantization and m-bit activation quantization. DDIM samplers \cite{ddpm} are adopted for LDMs and the PLMS sampler \cite{plms} is used for Stable Diffusion. We generate 256 samples per time step for constructing the calibration set. The Adam optimizer \cite{adam} is adopted and the learning rate for weight finetuning and scaling factor finetuning is set as $1e^{-5}$ and $1e^{-4}$ respectively. 

\begin{table}[t]
\small
\centering
\adjustbox{width=1.0\linewidth}{
\begin{tabular}{cccccc}
\toprule
\begin{tabular}[c]{@{}c@{}}Bit-width\\ (W/A)\end{tabular}     & Method   & \begin{tabular}[c]{@{}c@{}}Size\\ (MB)\end{tabular}   & FID$\downarrow$ & sFID$\downarrow$ & IS$\uparrow$     \\ \midrule
\textcolor{gray}{32/32}                                                     & \textcolor{gray}{FP}    &  \textcolor{gray}{1529.7}  & \textcolor{gray}{11.28} & \textcolor{gray}{7.70} & \textcolor{gray}{364.73} \\ \midrule
\multirow{3}{*}{8/8} & Q-Diffusion & 428.7 & 10.60 & 9.29 & 350.93 \\
                     & PTQ-D       & 428.7 & \bf{10.05} & 9.01 & 359.78 \\
                     & EfficientDM$^*$    & 435.0 & 11.38 & 8.04 & 362.34 \\
                     & \textbf{Ours}        & 428.7  & 10.43 & \bf{6.07} & \bf{365.12} \\ \midrule
\multirow{3}{*}{4/8} & Q-Diffusion & 237.5  & 9.29  & 9.29 & 336.80 \\
                     & PTQ-D       & 237.5  & 8.74  & 7.98 & 344.72 \\
                     & EfficientDM$^*$    & 243.8 & 9.93 & 7.34 & 353.83 \\
                     & \textbf{Ours}        & 237.5 & \bf{8.48}  & \bf{6.55} & \textbf{354.97} \\ \midrule
\multirow{3}{*}{4/4} & Q-Diffusion & 237.5  & N/A   & N/A  & N/A    \\
                     & PTQ-D       & 237.5  & N/A   & N/A  & N/A    \\
                     & EfficientDM$^*$    & 243.8 &  6.97 & 9.28 & 199.96 \\
                     & \textbf{Ours}        & 237.5  & \bf{5.98}  & \bf{7.93} & \bf{202.45} \\ \bottomrule
\end{tabular}
}
\vspace{-8pt}
\caption{Quantization performance on ImageNet 256$\times$256. ``*'' denotes the results obtained by re-running the open-source code.}
\label{tab:class_cond_result}
\vspace{-12pt}
\end{table}

We compare with popular PTQ methods including PTQ4DM \cite{ptq4dm}, Q-Diffusion \cite{qdiffusion} and PTQ-D \cite{ptqd}, as well as the state-of-the-art efficient finetuning method EfficientDM \cite{efficientdm}. 
The performance of different quantized LDMs is evaluated using the Fr$\acute{\text{e}}$chet Inception Distance (FID) \cite{fid}, spatial FID (sFID) \cite{sfid} and Inception Score (IS) \cite{is}. Unless specified, quantitative results are obtained by sampling 50,000 images and evaluated using the official evaluation scripts \cite{guided_diffusion}. For Stable Diffusion, we use the CLIP Score \cite{clipscore} for evaluation. All experiments are conducted on A6000 GPUs.


\subsection{Experiment Results and Analysis}
\label{sec:ablation}
\textbf{Unconditional Generation:} We evaluate the performance of our method over LDM-4 (LSUN-Bedrooms 256$\times$256) and LDM-8 (LSUN-Churches 256$\times$256) using the DDIM sampler with 200 and 500 time steps, respectively. Results are shown in \cref{tab:uncond_result} using FID, where our method outperforms the other baselines by a good margin. Note that the Inception Score is not a reasonable metric for datasets that have significantly different domains and categories from ImageNet \cite{qdiffusion}, thus not included. We further provide comparison with TFMQ-DM \cite{tfmq} in Appendix \ref{sec:tfmq_comp}.

\begin{table}
\small
\centering
\adjustbox{width=1.0\linewidth}{
\begin{tabular}{cccc}
\toprule
Bit-width (W/A)      & Method  & Size (MB)   & CLIP Score$\uparrow$ \\ \midrule
\textcolor{gray}{32/32}                & \textcolor{gray}{FP}          & \textcolor{gray}{3279.1} & \textcolor{gray}{31.50}      \\ \midrule
\multirow{2}{*}{8/8} & Q-Diffusion & 949.0  & 31.43      \\  
                     & \textbf{Ours}        & 949.0  & \bf{31.47} \\ \midrule
\multirow{2}{*}{4/8} & Q-Diffusion & 539.1  & 31.39      \\
                     & \textbf{Ours}        & 539.1  & \bf{31.50}      \\ \midrule
\multirow{2}{*}{4/4} & Q-Diffusion & 539.1  & N/A        \\
                     & \textbf{Ours}        & 539.1  & \bf{28.85}      \\ \bottomrule
\end{tabular}
}
\vspace{-8pt}
\caption{Quantization performance on Stable Diffusion v1.4 (512$\times$512) using COCO2014 prompts.}
\label{tab:sd_result}
\vspace{-16pt}
\end{table}

\begin{table*}[h!]
\small
\adjustbox{width=1.0\linewidth}{
\begin{minipage}{0.72\linewidth}
    \centering
    \adjustbox{width=1.0\linewidth}{
    \begin{tabular}{ccccccc}
    \toprule
    Method             & \begin{tabular}[c]{@{}c@{}}Bit-width \\ (W/A)\end{tabular} & \begin{tabular}[c]{@{}c@{}}Calibration \\ data size\end{tabular} & \begin{tabular}[c]{@{}c@{}}Time cost\\ (GPU hours)\end{tabular} & \begin{tabular}[c]{@{}c@{}}Memory cost\\ (MB)\end{tabular} & \begin{tabular}[c]{@{}c@{}}Model size\\ (MB)\end{tabular} & FID$\downarrow$   \\ \midrule
    \textcolor{gray}{FP}                 & \textcolor{gray}{32/32} & - & -     & -   & \textcolor{gray}{1045.6}  & \textcolor{gray}{2.95}  \\ \midrule
    PTQ \cite{qdiffusion} & 4/8  & 5120 & 23.08  & 10334 & 148.4 & 5.37   \\
    Baseline          & 4/8 & 5120 & 11.52   & 9822  & 148.4 & 6.95 \\
    + $\text{TLA}$ & 4/8  & 5120 & 13.13   & 11862 & 148.4 & 4.41  \\
    + $\text{TLA}$ + $\text{CMA}$   & 4/8  & 5120 & 15.25   & 12178 & 148.4 & \textbf{3.26} \\ \bottomrule
    \end{tabular}
    }
    \vspace{-6pt}
    \caption{Component and efficiency comparisons on LDM-4 (LSUN-Bedrooms 256 $\times$ 256). The baseline method is direct quantization with the Adaptive Rounding \cite{adaround} strategy.}
    \label{tab:component_comp}
    \vspace{-16pt}
\end{minipage}
\hspace{0.01\linewidth}  
\begin{minipage}{0.23\linewidth}
    \vspace{0pt}
    \adjustbox{width=1.0\linewidth}{
    \begin{tabular}{ccc}
    \toprule
    $\textbf{TLA}$ &  w/o $\mathcal{L}_{\text{G}}$ & w/ $\mathcal{L}_{\text{G}}$ \\ \midrule
    FID$\downarrow$       & 8.99 & 6.41 \\
    sFID$\downarrow$      & 15.23  & 11.18 \\ \midrule
    $\textbf{CMA}$  & w/o $\mathcal{L}_{\text{G}}$ & w/ $\mathcal{L}_{\text{G}}$ \\ \midrule
    FID$\downarrow$       & 11.41 & 6.20 \\
    sFID$\downarrow$      & 15.64 & 12.32 \\ \bottomrule
    \end{tabular}
    }
    \vspace{-6pt}
    \caption{Influence of global loss supervision on performance.}
    \label{tab:task_loss}
    \vspace{-16pt}
\end{minipage}
}
\end{table*}

\noindent\textbf{Class-conditional Generation:} We evaluate the performance using LDM-4 on ImageNet 256$\times$256 using the DDIM sampler (20 steps). As shown in \cref{tab:class_cond_result}, three metrics are used for evaluation. Note that sFID uses additional intermediate spatial features for calculation compared with FID. We can also see that FID is not a valid metric for ImageNet LDM-4 evaluation: All methods have lower FID when quantized to lower bits, conflicting with human perception. We show that our method not only succeeds in W4A4 quantization, but also improves the generation quality under higher bit settings. Under all three kinds of bit-width settings, our method is able to outperform the SOTA PTQ methods and EfficientDM in both sFID and IS. Examples of our generated images are included in Appendix \ref{sec:image_example}.


\noindent\textbf{Text-to-image Generation:} We use Stable Diffusion v1.4 as the model for quantization with the PLMS sampler sampling 50 time steps. \cref{tab:sd_result} shows the results. Images are generated based on the 10,000 prompts sampled from the COCO2014 \cite{coco} validation set, and CLIP Score is calculated based on the ViT-B/16 backbone. Given the limited works done on Stable Diffusion, we can only compare with Q-Diffusion and the full-precision baseline.

\begin{figure}[h]
\vspace{-12pt}
\centering
\centerline{\includegraphics[trim=380 400 690 230,clip,width=1\linewidth]{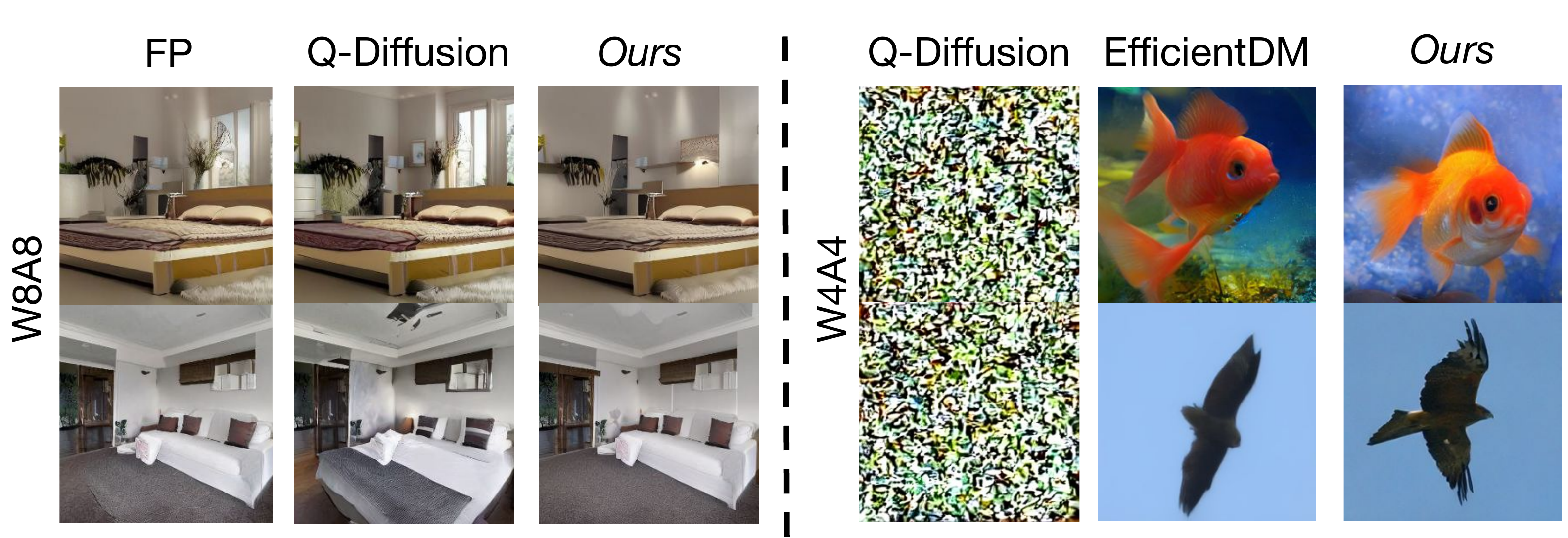}}
\vspace{-10pt}
\caption{Visual comparison with Q-Diffusion and EfficientDM. QuEST outperforms the baselines with better visual quality.}
\label{fig:example_image}
\vspace{-12pt}
\end{figure}

\subsection{Ablations and Discussions}
\noindent\textbf{Efficiency comparison with PTQ methods and the impact of individual components.}
\cref{tab:component_comp} compares the efficiency and performance against the post-training quantization (PTQ) approach on the LSUN-Bedrooms dataset. Although our method uses the same amount of calibration data as the PTQ approach, it achieves better time efficiency with only a 20\% increase in GPU memory usage. We also illustrate the contribution of each component to generation performance. The results indicate that sequentially finetuning the time embedding layers, followed by attention-related layers, yields consistent performance improvements.

\cref{tab:task_loss} presents a comparison of performance with and without the global loss $\mathcal{L}_{\text{G}}$. The results indicate that supervising the quantized model using the output difference from the full-precision counterpart is essential for performance improvement, enhancing the FID by 2.58 and 5.21 for TLA and CMA, respectively. However, when the learning process is only supervised by the global loss, we find that the performance degrades by 7.13 FID and 9.39 sFID for TLA, suggesting that the global loss alone is insufficient for optimal performance.


\begin{table}[h]
\small
\centering
\vspace{-8pt}
\adjustbox{width=1.0\linewidth}{
\begin{tabular}{cc|cccc}
\toprule
Bit-width& Method  & \begin{tabular}[c]{@{}c@{}}Time \\ (h)\end{tabular} & \begin{tabular}[c]{@{}c@{}}Memory\\ (MB)\end{tabular} & Iters  & FID $\downarrow$  \\ \midrule
& EfficientDM   & 2.60  & 12004  & 32k  & 15.15       \\
& Full-finetune & 0.85  & 15076  & 2.2k & 5.38         \\
\multirow{-3}{*}{W4A8} & \textbf{Ours} & 0.45  & 12178  & 2.2k & \textbf{3.82} \\ \midrule
& EfficientDM   & 2.60  & 12004  & 32k  & 10.60    \\
& Full-finetune & 0.85  & 15076 & 2.2k &  6.36  \\
\multirow{-3}{*}{W4A4} & \textbf{Ours} & 0.45  & 12178   & 2.2k & \textbf{6.12} \\ \bottomrule
\end{tabular}
}
\vspace{-8pt}
\caption{Efficiency comparison with other finetuning methods.}
\label{tab:effciency}
\vspace{-8pt}
\end{table}

\noindent\textbf{How QuEST adjusts the activation distribution.} Our approach is motivated by the imbalanced activation distribution in diffusion models, hence we aim to analyze how our fine-tuning strategy addresses this challenge.
As shown in \cref{fig:activation_range}, our method refines the activation distribution, making it more conducive to quantization. Specifically, the activation value ranges shrink from [-10, 34] to [-4, 14] and from [-11, 20] to [-4, 4]. Additionally, the standard deviations decrease from 0.171 to 0.157 and from 0.073 to 0.071, while the mean remains consistent. This results in a more compact activation distribution, effectively reducing both rounding and clipping errors during quantization.

\noindent\textbf{Comparison with precomputed time embeddings.} In diffusion models, time embeddings are independent of input conditions and noise. A potential approach is to precompute these embeddings and reuse them directly. However, this strategy overlooks the compatibility between different modules in a quantized model. We take this into consideration and optimize the time embeddings with $\arg\min_{\mathbf{w}_l} (\mathcal{L}_{\text{TLA}} + \mathcal{L}_{\text{G}}), \ \ l \in \mathbb{C}_{\text{TE}}$ so that the time embedding layers are also trained to minimize the final prediction error. 
As shown in \cref{tab:time_embed}, adding this optimization objective enhances quantization performance, even surpassing the full-precision baseline (which uses precomputed features).

\noindent\textbf{Integration with LoRA finetuning.} Different ways exist for finetuning quantized models. We further employ QALoRA \cite{efficientdm} to finetune on the ImageNet 256$\times$256 dataset. A rank of 32 is used for the LoRA weights, and the parameters are trained over 100 time steps for 160 epochs. We find that integrating the QALoRA technique leads to a 5.62 increase in FID, indicating that finetuning the original layers is a better solution for performance preservation.

\noindent\textbf{Efficiency comparison with other finetuning methods.} 
We compare with EfficientDM and full-finetuning in terms of actual training costs on LDM-4 in \cref{tab:effciency}. The setting of full-finetuning is aligned with our method. We observe that: compared with EfficientDM, our method requires fewer training iterations and time to obtain better performance with comparable GPU memory cost. Compared with full-finetuning, our method costs less time and memory, as well as achieving better performance. The bottleneck in computational costs becomes more severe when scaled to larger models such as Stable Diffusion. We find that while full-finetuning quickly encounters OOM, our method is able to finetune SD on a single GPU with 48GB memory.

\section{Conclusion}
We have proposed QuEST, an efficient data-free finetuning framework for low-bit diffusion model quantization. Our method is motivated by the current challenge in low-bit diffusion model quantization and guided by the two underlying properties found in quantized diffusion models. 
To alleviate the performance degradation, we propose to finetune the time embedding layers and the attention-related layers under the supervision of the full-precision counterpart.
Experimental results on three high-resolution image generation tasks (including Stable Diffusion) demonstrate the effectiveness and efficiency of QuEST, achieving low-bit compatibility with less time and memory cost. 

\noindent \textbf{Acknowledgments:} This research is supported by NSF IIS-2525840, CNS-2432534, ECCS-2514574, NIH 1RF1MH133764-01 and Cisco Research unrestricted gift. This article solely reflects opinions and conclusions of authors and not funding agencies.

{
    \small
    \bibliographystyle{ieeenat_fullname}
    \bibliography{main}

\begin{thebibliography}{41}
\providecommand{\natexlab}[1]{#1}
\providecommand{\url}[1]{\texttt{#1}}
\expandafter\ifx\csname urlstyle\endcsname\relax
  \providecommand{\doi}[1]{doi: #1}\else
  \providecommand{\doi}{doi: \begingroup \urlstyle{rm}\Url}\fi

\bibitem[Barratt and Sharma(2018)]{is}
Shane Barratt and Rishi Sharma.
\newblock A note on the inception score, 2018.

\bibitem[Choi et~al.(2022)Choi, Lee, Shin, Kim, Kim, and Yoon]{perceptiondm}
Jooyoung Choi, Jungbeom Lee, Chaehun Shin, Sungwon Kim, Hyunwoo Kim, and Sungroh Yoon.
\newblock Perception prioritized training of diffusion models, 2022.

\bibitem[Croitoru et~al.(2023)Croitoru, Hondru, Ionescu, and Shah]{diffusion_survey}
Florinel-Alin Croitoru, Vlad Hondru, Radu~Tudor Ionescu, and Mubarak Shah.
\newblock Diffusion models in vision: A survey.
\newblock \emph{IEEE Transactions on Pattern Analysis and Machine Intelligence}, 45\penalty0 (9):\penalty0 10850--10869, 2023.

\bibitem[Deng et~al.(2009)Deng, Dong, Socher, Li, Li, and Fei-Fei]{imagenet}
Jia Deng, Wei Dong, Richard Socher, Li-Jia Li, Kai Li, and Li Fei-Fei.
\newblock Imagenet: A large-scale hierarchical image database.
\newblock In \emph{2009 IEEE conference on computer vision and pattern recognition}, pages 248--255. IEEE, 2009.

\bibitem[Dettmers et~al.(2023)Dettmers, Pagnoni, Holtzman, and Zettlemoyer]{qlora}
Tim Dettmers, Artidoro Pagnoni, Ari Holtzman, and Luke Zettlemoyer.
\newblock Qlora: Efficient finetuning of quantized llms, 2023.

\bibitem[Dhariwal and Nichol(2021)]{guided_diffusion}
Prafulla Dhariwal and Alex Nichol.
\newblock Diffusion models beat gans on image synthesis, 2021.

\bibitem[Feng et~al.(2023)Feng, Zhang, Yu, Fang, Li, Chen, Lu, Liu, Yin, Feng, Sun, Chen, Tian, Wu, and Wang]{ffl}
Zhida Feng, Zhenyu Zhang, Xintong Yu, Yewei Fang, Lanxin Li, Xuyi Chen, Yuxiang Lu, Jiaxiang Liu, Weichong Yin, Shikun Feng, Yu Sun, Li Chen, Hao Tian, Hua Wu, and Haifeng Wang.
\newblock Ernie-vilg 2.0: Improving text-to-image diffusion model with knowledge-enhanced mixture-of-denoising-experts.
\newblock In \emph{Proceedings of the IEEE/CVF Conference on Computer Vision and Pattern Recognition (CVPR)}, pages 10135--10145, 2023.

\bibitem[Gholami et~al.(2022)Gholami, Kim, Dong, Yao, Mahoney, and Keutzer]{quantization_survey}
Amir Gholami, Sehoon Kim, Zhen Dong, Zhewei Yao, Michael~W Mahoney, and Kurt Keutzer.
\newblock A survey of quantization methods for efficient neural network inference.
\newblock In \emph{Low-Power Computer Vision}, pages 291--326. Chapman and Hall/CRC, 2022.

\bibitem[Guo et~al.(2023)Guo, Lu, Liu, Cheng, and Hu]{van}
Meng-Hao Guo, Cheng-Ze Lu, Zheng-Ning Liu, Ming-Ming Cheng, and Shi-Min Hu.
\newblock Visual attention network.
\newblock \emph{Computational Visual Media}, 9\penalty0 (4):\penalty0 733--752, 2023.

\bibitem[He et~al.(2023{\natexlab{a}})He, Liu, Wu, Zhou, and Zhuang]{efficientdm}
Yefei He, Jing Liu, Weijia Wu, Hong Zhou, and Bohan Zhuang.
\newblock Efficientdm: Efficient quantization-aware fine-tuning of low-bit diffusion models, 2023{\natexlab{a}}.

\bibitem[He et~al.(2023{\natexlab{b}})He, Liu, Liu, Wu, Zhou, and Zhuang]{ptqd}
Yefei He, Luping Liu, Jing Liu, Weijia Wu, Hong Zhou, and Bohan Zhuang.
\newblock Ptqd: Accurate post-training quantization for diffusion models, 2023{\natexlab{b}}.

\bibitem[Hessel et~al.(2022)Hessel, Holtzman, Forbes, Bras, and Choi]{clipscore}
Jack Hessel, Ari Holtzman, Maxwell Forbes, Ronan~Le Bras, and Yejin Choi.
\newblock Clipscore: A reference-free evaluation metric for image captioning, 2022.

\bibitem[Heusel et~al.(2018)Heusel, Ramsauer, Unterthiner, Nessler, and Hochreiter]{fid}
Martin Heusel, Hubert Ramsauer, Thomas Unterthiner, Bernhard Nessler, and Sepp Hochreiter.
\newblock Gans trained by a two time-scale update rule converge to a local nash equilibrium, 2018.

\bibitem[Ho et~al.(2020)Ho, Jain, and Abbeel]{ddpm}
Jonathan Ho, Ajay Jain, and Pieter Abbeel.
\newblock Denoising diffusion probabilistic models, 2020.

\bibitem[Hu et~al.(2021)Hu, Shen, Wallis, Allen-Zhu, Li, Wang, Wang, and Chen]{lora}
Edward~J. Hu, Yelong Shen, Phillip Wallis, Zeyuan Allen-Zhu, Yuanzhi Li, Shean Wang, Lu Wang, and Weizhu Chen.
\newblock Lora: Low-rank adaptation of large language models, 2021.

\bibitem[Huang et~al.(2024)Huang, Gong, Liu, Chen, and Liu]{tfmq}
Yushi Huang, Ruihao Gong, Jing Liu, Tianlong Chen, and Xianglong Liu.
\newblock Tfmq-dm: Temporal feature maintenance quantization for diffusion models, 2024.

\bibitem[Jacob et~al.(2017)Jacob, Kligys, Chen, Zhu, Tang, Howard, Adam, and Kalenichenko]{qat}
Benoit Jacob, Skirmantas Kligys, Bo Chen, Menglong Zhu, Matthew Tang, Andrew Howard, Hartwig Adam, and Dmitry Kalenichenko.
\newblock Quantization and training of neural networks for efficient integer-arithmetic-only inference, 2017.

\bibitem[Kingma and Ba(2017)]{adam}
Diederik~P. Kingma and Jimmy Ba.
\newblock Adam: A method for stochastic optimization, 2017.

\bibitem[Li et~al.(2023{\natexlab{a}})Li, Liu, Lian, Yang, Dong, Kang, Zhang, and Keutzer]{qdiffusion}
Xiuyu Li, Yijiang Liu, Long Lian, Huanrui Yang, Zhen Dong, Daniel Kang, Shanghang Zhang, and Kurt Keutzer.
\newblock Q-diffusion: Quantizing diffusion models.
\newblock In \emph{Proceedings of the IEEE/CVF International Conference on Computer Vision (ICCV)}, pages 17535--17545, 2023{\natexlab{a}}.

\bibitem[Li et~al.(2021)Li, Gong, Tan, Yang, Hu, Zhang, Yu, Wang, and Gu]{brecq}
Yuhang Li, Ruihao Gong, Xu Tan, Yang Yang, Peng Hu, Qi Zhang, Fengwei Yu, Wei Wang, and Shi Gu.
\newblock Brecq: Pushing the limit of post-training quantization by block reconstruction.
\newblock \emph{arXiv preprint arXiv:2102.05426}, 2021.

\bibitem[Li et~al.(2022)Li, Xu, Zhang, Cao, Gao, and Guo]{qvit}
Yanjing Li, Sheng Xu, Baochang Zhang, Xianbin Cao, Peng Gao, and Guodong Guo.
\newblock Q-vit: Accurate and fully quantized low-bit vision transformer, 2022.

\bibitem[Li et~al.(2023{\natexlab{b}})Li, Xu, Cao, Zhang, and Sun]{qdm}
Yanjing Li, Sheng Xu, Xianbin Cao, Baochang Zhang, and Xiao Sun.
\newblock Q-dm: An efficient low-bit quantized diffusion model.
\newblock In \emph{NeurIPS 2023}, 2023{\natexlab{b}}.

\bibitem[Liang et~al.(2021)Liang, Glossner, Wang, Shi, and Zhang]{quantization_survey2}
Tailin Liang, John Glossner, Lei Wang, Shaobo Shi, and Xiaotong Zhang.
\newblock Pruning and quantization for deep neural network acceleration: A survey.
\newblock \emph{Neurocomputing}, 461:\penalty0 370--403, 2021.

\bibitem[Lin et~al.(2015)Lin, Maire, Belongie, Bourdev, Girshick, Hays, Perona, Ramanan, Zitnick, and Dollár]{coco}
Tsung-Yi Lin, Michael Maire, Serge Belongie, Lubomir Bourdev, Ross Girshick, James Hays, Pietro Perona, Deva Ramanan, C.~Lawrence Zitnick, and Piotr Dollár.
\newblock Microsoft coco: Common objects in context, 2015.

\bibitem[Liu et~al.(2023)Liu, Niu, Yuan, Yang, Wang, and Liu]{pdquant}
Jiawei Liu, Lin Niu, Zhihang Yuan, Dawei Yang, Xinggang Wang, and Wenyu Liu.
\newblock Pd-quant: Post-training quantization based on prediction difference metric, 2023.

\bibitem[Liu et~al.(2022)Liu, Ren, Lin, and Zhao]{plms}
Luping Liu, Yi Ren, Zhijie Lin, and Zhou Zhao.
\newblock Pseudo numerical methods for diffusion models on manifolds, 2022.

\bibitem[Ma et~al.(2024)Ma, Yang, and Huang]{diff_img_trans}
Hao Ma, Jingyuan Yang, and Hui Huang.
\newblock Taming diffusion model for exemplar-based image translation.
\newblock \emph{Computational Visual Media}, 10\penalty0 (6):\penalty0 1031--1043, 2024.

\bibitem[Nagel et~al.(2020)Nagel, Amjad, Van~Baalen, Louizos, and Blankevoort]{adaround}
Markus Nagel, Rana~Ali Amjad, Mart Van~Baalen, Christos Louizos, and Tijmen Blankevoort.
\newblock Up or down? adaptive rounding for post-training quantization.
\newblock In \emph{International Conference on Machine Learning}, pages 7197--7206. PMLR, 2020.

\bibitem[Nash et~al.(2021)Nash, Menick, Dieleman, and Battaglia]{sfid}
Charlie Nash, Jacob Menick, Sander Dieleman, and Peter~W. Battaglia.
\newblock Generating images with sparse representations, 2021.

\bibitem[Rombach et~al.(2021)Rombach, Blattmann, Lorenz, Esser, and Ommer]{ldm}
Robin Rombach, Andreas Blattmann, Dominik Lorenz, Patrick Esser, and Björn Ommer.
\newblock High-resolution image synthesis with latent diffusion models, 2021.

\bibitem[Ronneberger et~al.(2015)Ronneberger, Fischer, and Brox]{unet}
Olaf Ronneberger, Philipp Fischer, and Thomas Brox.
\newblock U-net: Convolutional networks for biomedical image segmentation, 2015.

\bibitem[Shang et~al.(2023)Shang, Yuan, Xie, Wu, and Yan]{ptq4dm}
Yuzhang Shang, Zhihang Yuan, Bin Xie, Bingzhe Wu, and Yan Yan.
\newblock Post-training quantization on diffusion models.
\newblock In \emph{CVPR}, 2023.

\bibitem[So et~al.(2023)So, Lee, Ahn, Kim, and Park]{tdqd}
Junhyuk So, Jungwon Lee, Daehyun Ahn, Hyungjun Kim, and Eunhyeok Park.
\newblock Temporal dynamic quantization for diffusion models, 2023.

\bibitem[Sui et~al.(2024)Sui, Li, Kag, Idelbayev, Cao, Hu, Sagar, Yuan, Tulyakov, and Ren]{bitfusion}
Yang Sui, Yanyu Li, Anil Kag, Yerlan Idelbayev, Junli Cao, Ju Hu, Dhritiman Sagar, Bo Yuan, Sergey Tulyakov, and Jian Ren.
\newblock Bitsfusion: 1.99 bits weight quantization of diffusion model, 2024.

\bibitem[Wang et~al.(2023)Wang, Wang, Xu, Tang, Zhou, and Lu]{adp-dm}
Changyuan Wang, Ziwei Wang, Xiuwei Xu, Yansong Tang, Jie Zhou, and Jiwen Lu.
\newblock Towards accurate data-free quantization for diffusion models, 2023.

\bibitem[Wang et~al.(2025)Wang, Peng, Liu, Gu, and Hu]{diffusion_survey2}
Chen Wang, Hao-Yang Peng, Ying-Tian Liu, Jiatao Gu, and Shi-Min Hu.
\newblock Diffusion models for 3d generation: A survey.
\newblock \emph{Computational Visual Media}, 11\penalty0 (1):\penalty0 1--28, 2025.

\bibitem[Wei et~al.(2023)Wei, Gong, Li, Liu, and Yu]{qdrop}
Xiuying Wei, Ruihao Gong, Yuhang Li, Xianglong Liu, and Fengwei Yu.
\newblock Qdrop: Randomly dropping quantization for extremely low-bit post-training quantization, 2023.

\bibitem[Wu et~al.(2024)Wu, Wang, Shang, Shah, and Yan]{ptq4dit}
Junyi Wu, Haoxuan Wang, Yuzhang Shang, Mubarak Shah, and Yan Yan.
\newblock Ptq4dit: Post-training quantization for diffusion transformers, 2024.

\bibitem[Yang et~al.(2023)Yang, Dai, Wang, Zhang, and Zhang]{eqs}
Yuewei Yang, Xiaoliang Dai, Jialiang Wang, Peizhao Zhang, and Hongbo Zhang.
\newblock Efficient quantization strategies for latent diffusion models, 2023.

\bibitem[Yu et~al.(2015)Yu, Zhang, Song, Seff, and Xiao]{lsun}
Fisher Yu, Yinda Zhang, Shuran Song, Ari Seff, and Jianxiong Xiao.
\newblock Lsun: Construction of a large-scale image dataset using deep learning with humans in the loop.
\newblock \emph{arXiv preprint arXiv:1506.03365}, 2015.

\bibitem[Yuan et~al.(2022)Yuan, Xue, Chen, Wu, and Sun]{ptq4vit}
Zhihang Yuan, Chenhao Xue, Yiqi Chen, Qiang Wu, and Guangyu Sun.
\newblock Ptq4vit: Post-training quantization framework for vision transformers with twin uniform quantization, 2022.

\end{thebibliography}
}

\clearpage
\setcounter{page}{1}
\maketitlesupplementary

The supplementary material is organized as follows: \cref{sec:tfmq_comp} provides comparison with TFMQ-DM; \cref{sec:low_res} provides comparison on the low-resolution dataset; \cref{sec:proof} provides the proof and detailed analysis for \cref{th:weight_finetune}; \cref{sec:example_dist} presents additional examples of the imbalanced distributions across different models; \cref{sec:large_act} highlights the importance of the large values in activations; \cref{sec:image_example} offers further generated examples from our method across varying bit-widths; and \cref{sec:limitations} discusses limitations and broader considerations.

\section{More Baseline Comparisons}
\label{sec:tfmq_comp}
We further compare with TFMQ \cite{tfmq} below:
\begin{table}[h]
    \centering
    \vspace{-4pt}
    \small
    \adjustbox{width=0.6\linewidth}{
        \begin{tabular}{c|cc}
        \toprule
        Bedroom & W8A8 & W4A8 \\
        \midrule
          TFMQ-DM & 3.14 & 3.68\\
          QuEST & \textbf{3.03} & \textbf{3.26} \\
          \midrule
          ImageNet & W8A8 & W4A8 \\ 
          \midrule
          TFMQ-DM & 10.79 & 10.29 \\
          QuEST & \textbf{10.43} & \textbf{8.48} \\
          \bottomrule
        \end{tabular}
    }
    \vspace{-8pt}
    \caption{Comparing TFMQ.}
    \label{tab:tfmq}
    \vspace{-12pt}
\end{table}

We also supplement the metrics for Table \ref{tab:uncond_result}:
\begin{table}[h]
    \centering
    \vspace{-4pt}
    \small
    \adjustbox{width=0.6\linewidth}{
        \begin{tabular}{c|cc}
        \toprule
        \textit{W8A8} & sFID $\downarrow$ & IS $\uparrow$ \\ \midrule
        QDiffusion & 8.19 & 2.25 \\ 
        PTQD & 9.89 & 2.25 \\ 
        EfficientDM & N/A & N/A \\
        \textbf{Ours} & \textbf{6.86} & \textbf{2.27} \\ \midrule
        \textit{W4A4} & sFID $\downarrow$ & IS $\uparrow$ \\ \midrule
        QDiffusion & N/A & N/A \\ 
        PTQD & N/A & N/A \\ 
        EfficientDM & 15.15 & \textbf{2.27} \\
        \textbf{Ours} & \textbf{7.82} & 2.26 \\
    \bottomrule
    \end{tabular}
    }
    \vspace{-8pt}
    \caption{Additional metrics on LSUN-Bedrooms. ``N/A'' represents generation failure.}
    \label{tab:add_metric}
    \vspace{-15pt}
\end{table}

\section{Low-resolution dataset comparison}
\label{sec:low_res}
We further include experiments on CIFAR10 in \cref{tab:cifar10}.
\begin{table}[h]
    \centering
    \vspace{-8pt}
    \adjustbox{width=0.6\linewidth}{
        \begin{tabular}{c|cc}
        \toprule
             & W8A8 & W4A4 \\
        \midrule
            Q-Diffusion & 3.75 & N/A \\
            EfficientDM & 3.75 & 10.48 \\
            QuEST & \textbf{3.71} & \textbf{9.37} \\
        \bottomrule
        \end{tabular}
    }
    \vspace{-8pt}
    \caption{FID comparison on CIFAR10.}
    \label{tab:cifar10}
    \vspace{-12pt}
\end{table}

\section{Proof for \cref{th:weight_finetune}}
\label{sec:proof}
We provide the detailed proof for \cref{th:weight_finetune} here. The notations are consistent with the ones in the main paper. 

Since the perturbation $\Delta$ is too large for accurate Taylor expansion, we can resolve it by introducing a new perturbation $\epsilon=\Delta / K$, where we divide $\Delta$ by a constant $K$ so that $\epsilon$ is small enough for approximation. Then, \cref{eq:quant_err} is rewritten as follows:
\begin{align}
\label{eq:quant_err_new}
    &\mathbb{E}[L(z_{n,t} + \Delta; \mathbf{w})] - \mathbb{E}[L(z_{n,t}; \mathbf{w})]  \notag \\
    =& \sum_{i=1}^{K} \left (\mathbb{E}[L(z_{n,t} + \frac{i}{K}\Delta; \mathbf{w})] - \mathbb{E}[L(z_{n,t} + \frac{i-1}{K}\Delta; \mathbf{w})] \right) \notag \\
    \approx & \sum_{i=1}^{K} \left (\epsilon^{T}\overline{\mathbf{g}}^{(z_{n,t}+(i-1)\epsilon)} 
+ \frac{1}{2}\epsilon^{T}\overline{\mathbf{H}}^{(z_{n,t}+(i-1)\epsilon)}\epsilon \right),
\end{align}
where the approximation step follows Taylor expansion and only the first two main components are kept. The first term in \cref{eq:quant_err_new} cannot be ignored because samples such as $z_{n,t}+(i-1)\epsilon$ may not be included in the learned distribution of the model. The second term can still be minimized by reconstruction since only the difference between quantized model output and ground-truth matters. In the following, we temporarily exclude the second term for simplicity since it can always be minimized through aligning the activation outputs.

\begin{figure}[]
\begin{center}
\includegraphics[trim=30 75 50 60,clip,width=0.8\columnwidth]{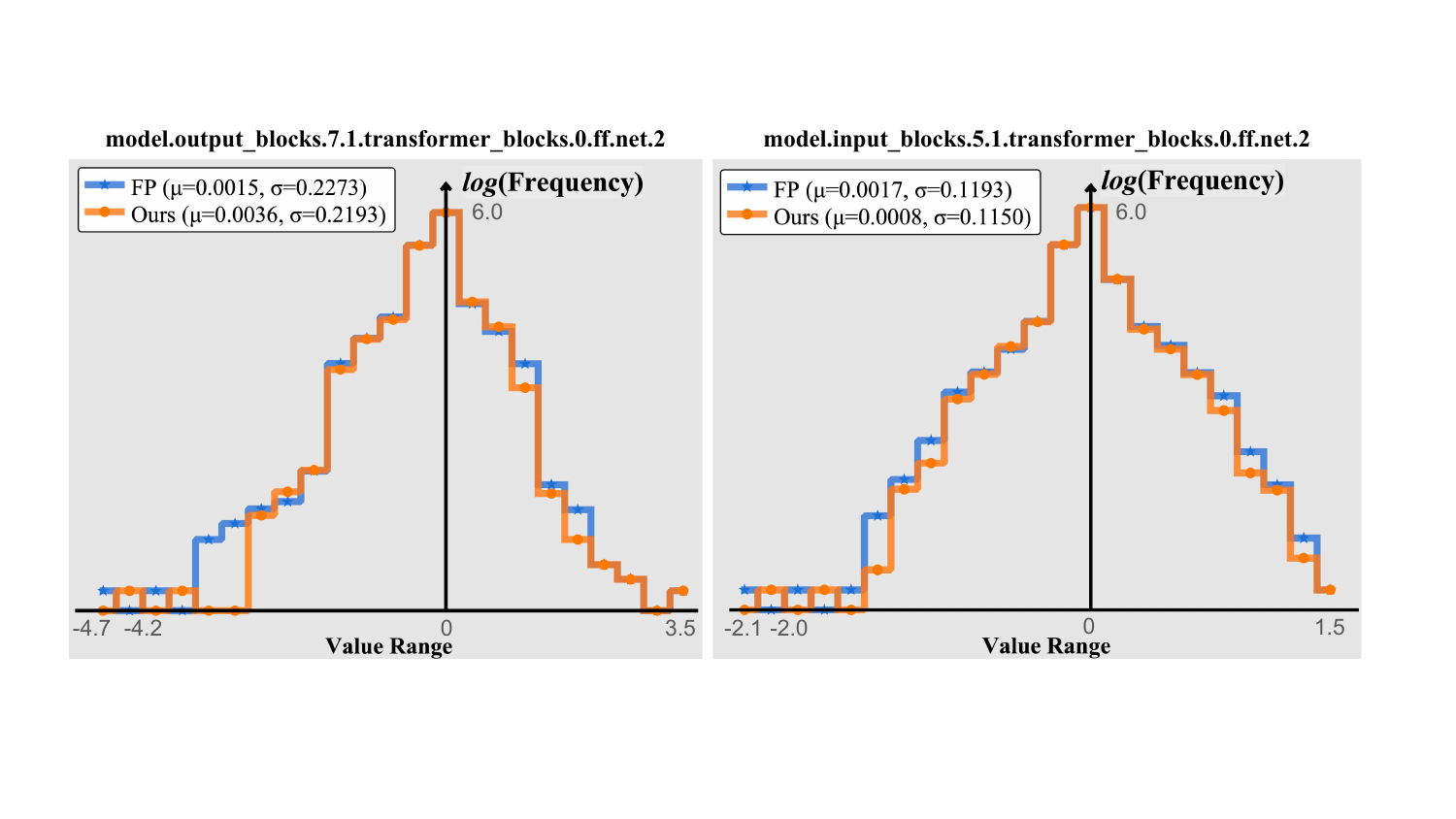} \\ 
\vspace{-0.05in}
(a) Activation Distribution on Conditional LDM4 (ImageNet 256 $\times$ 256) \\
\vspace{0.05in}
\quad \includegraphics[trim=30 75 30 80,clip,width=0.85\columnwidth]{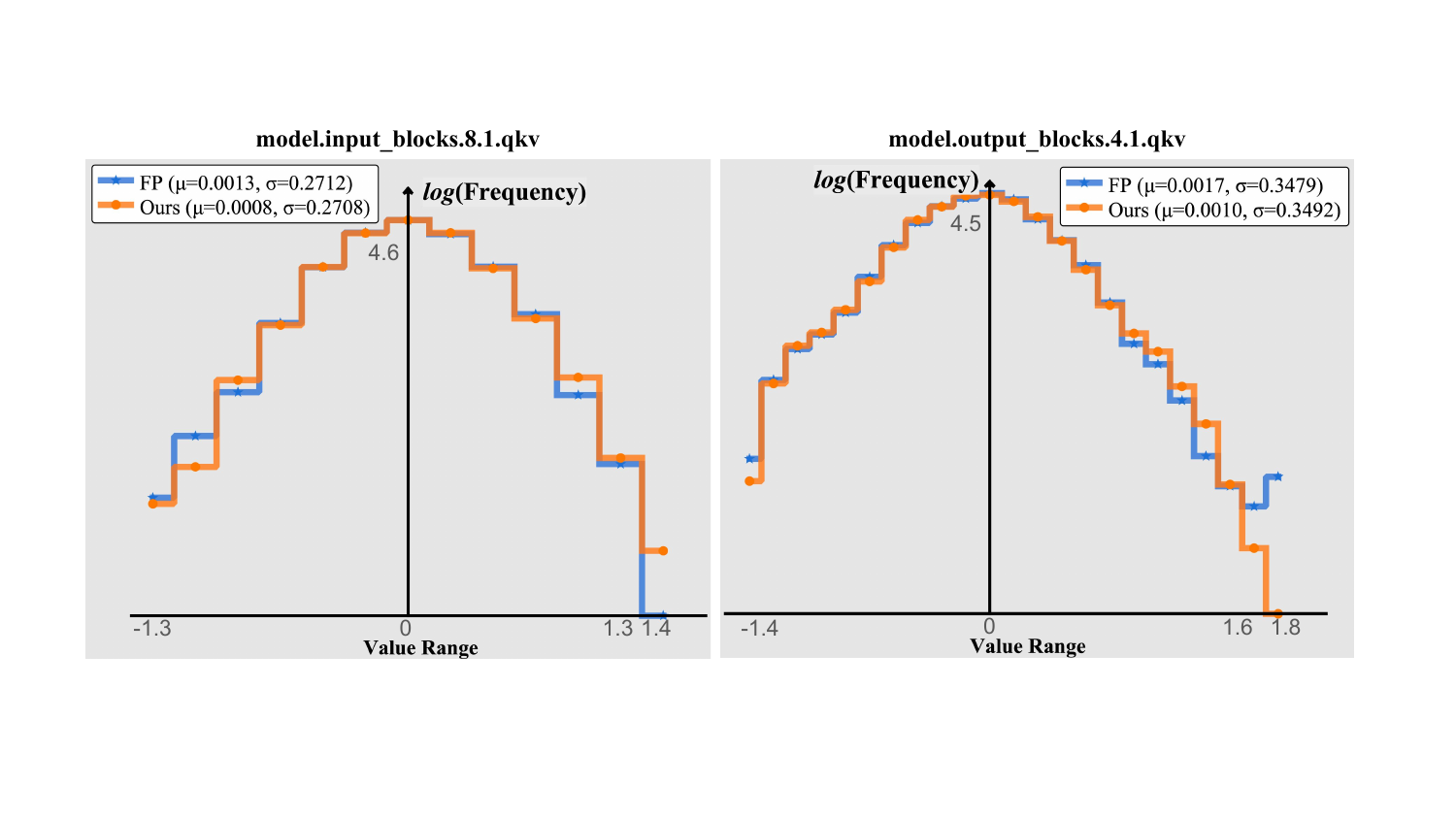} \\
\vspace{-0.05in}
(b) Activation Distribution on Unconditional LDM4 (LSUN-Bedrooms 256 $\times$ 256)\\
\vspace{-0.05in}
\caption{Illustrations of imbalanced activation distributions on conditional LDM4 (ImageNet 256$\times$256) and unconditional LDM4 (LSUN-Bedrooms 256$\times$256). }
\label{fig:appendix_act}
\end{center}
\vspace{-12pt}
\end{figure}

\begin{figure*}[h]
    \centering
    \includegraphics[trim=100 10 100 20,clip,width=0.8\textwidth]{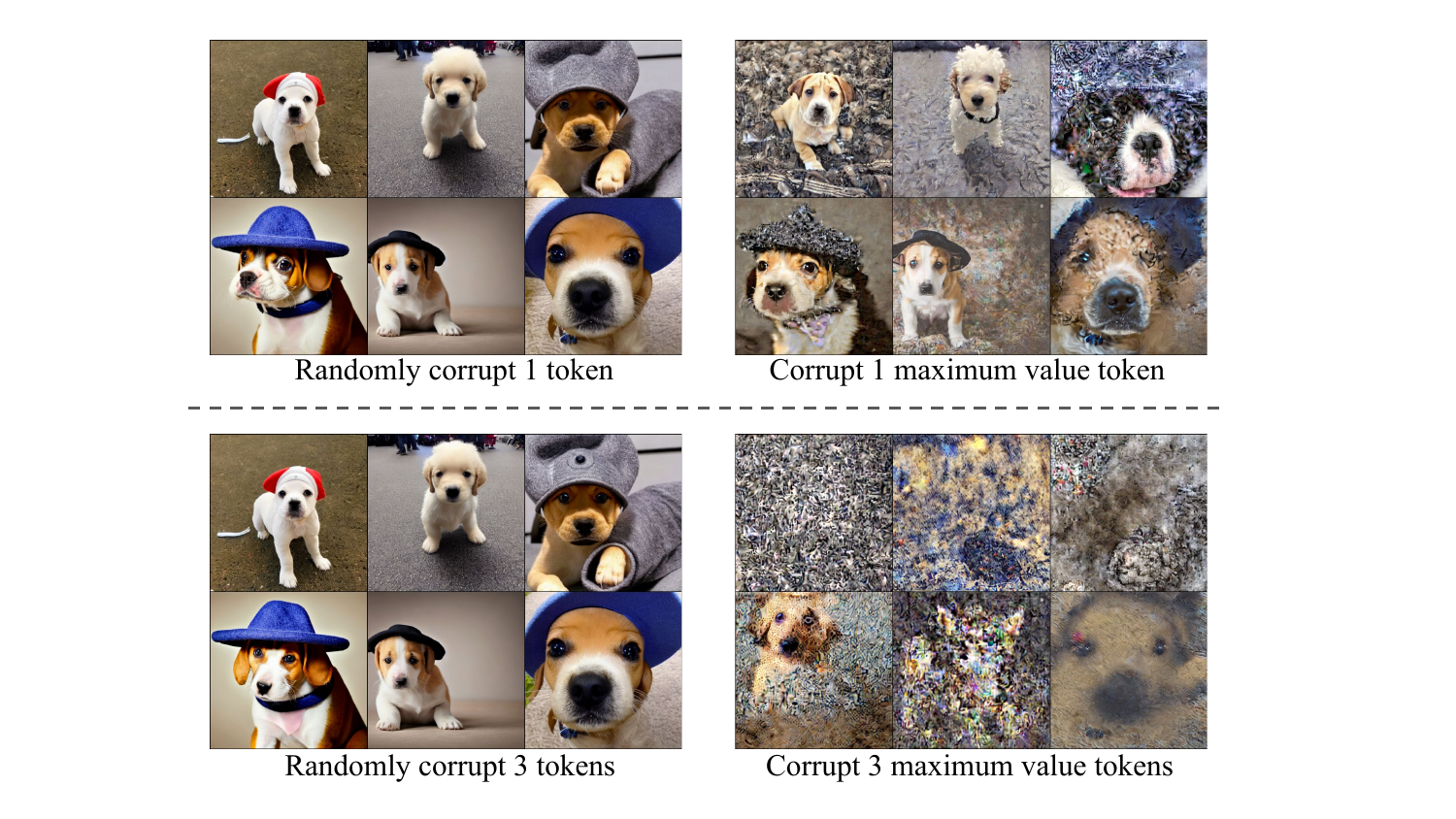}
    \vspace{-0.15in}
    \caption{Comparison of different corruptions made on different tokens.}
    \label{fig:corrupt_compare}
\end{figure*}

Given the objective function (MSE loss) of diffusion models, we analyze that:
\begin{align}
\label{eq:grad_expand}
    \sum_{i=1}^{K} \epsilon^{T}\overline{\mathbf{g}}^{(z_{n,t}+(i-1)\epsilon)} & = 2\epsilon^{T} \sum_{i=1}^{K} (\tilde{z}_{n-1,t}^{i} \cdot \mathbf{w}_{n} - \overline{z}_{n,t})  \notag \\
    \approx & 2\epsilon^{T} \sum_{i=1}^{K} (\tilde{z}_{n-1,t}^{i} \cdot \mathbf{w}_{n} - z_{\text{FP}}),
\end{align}
where $\mathbf{w}_{n}$ is the weight for layer $n$, $\tilde{z}_{n-1,t}^{i}$ is the activation of the $(n-1)$th layer in a quantized model to get $z_{n,t}+(i-1)\epsilon$. Ground-truth $\overline{z}_{n,t}$ can be approximated by the full-precision output $z_{\text{FP}}$. We see that $\tilde{z}_{n-1,t}^{i}$ and $z_{\text{FP}}$ cannot be changed, thus to minimize \cref{eq:grad_expand}, we need to finetune $\textbf{w}_{n}$. From a general perspective, \cref{eq:grad_expand} also indicates that the model has not converged well to a local minimum given the perturbed inputs, thus when we finetune the model layers given the quantized inputs, we are actually training the model towards convergence over new samples and increasing its robustness.

\section{Examples of Imbalanced Activation Distributions}
\label{sec:example_dist}
Apart from \cref{fig:activation_range}, we show that the imbalance in the activation distribution is a common phenomenon in different model structures and datasets. In \cref{fig:appendix_act}, we show more results of activation distributions of latent diffusion models on ImageNet 256 $\times$ 256 and LSUN-Bedrooms 256 $\times$ 256.

\section{Importance of large values in activations}
\label{sec:large_act}
As shown in \cref{fig:activation_range}, quite a few values are rather large and diversely distributed. These values pose difficulties on activation quantization, and being rather important and not negligible. To demonstrate this, we corrupt certain tokens in the activation outputs of the diffusion model and check the corresponding generated images. The corruption is done by setting the token values as all zeros. As shown in \cref{fig:corrupt_compare}, we compare two settings: (1) corrupt a certain number of tokens randomly; (2) corrupt the same number of the tokens with the largest values. 

\begin{figure}[h]
    \centering
    \begin{subfigure}[b]{0.48\textwidth}
    \includegraphics[trim=150 150 120 20,clip,width=\textwidth]{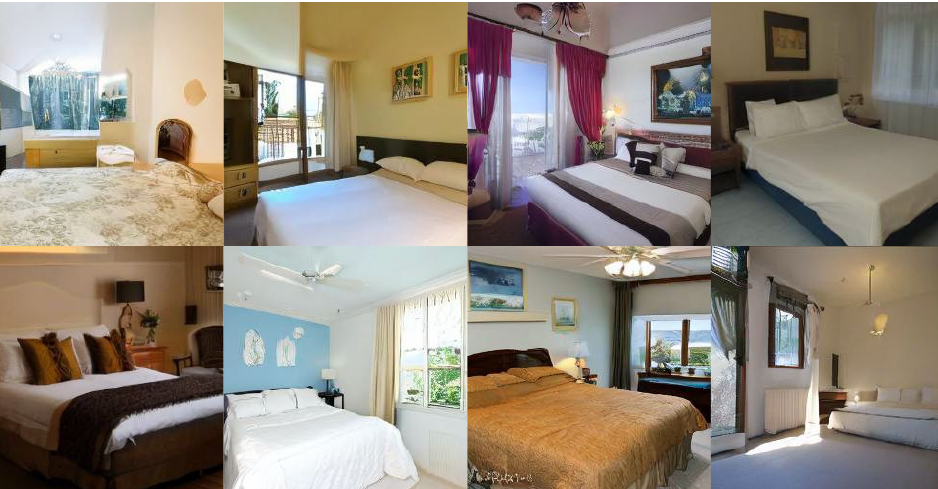}
    \vspace{-0.2in}
        \caption{Full Precision}
    \end{subfigure}
    \hfill
    \begin{subfigure}[b]{0.48\textwidth}
    \includegraphics[trim=150 150 120 20,clip,width=\textwidth]{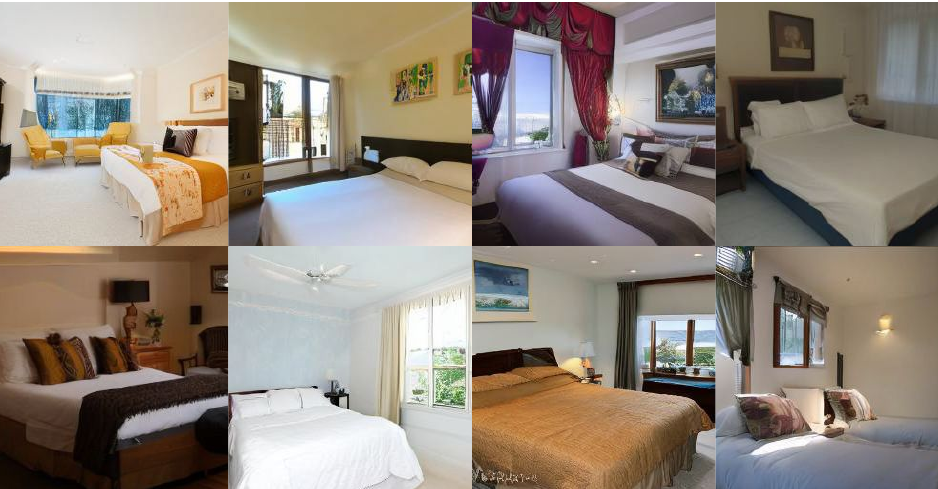}
    \vspace{-0.2in}
        \caption{W8A8}
    \end{subfigure} \\
    \begin{subfigure}[b]{0.48\textwidth}
    \includegraphics[trim=150 150 120 20,clip,width=\textwidth]{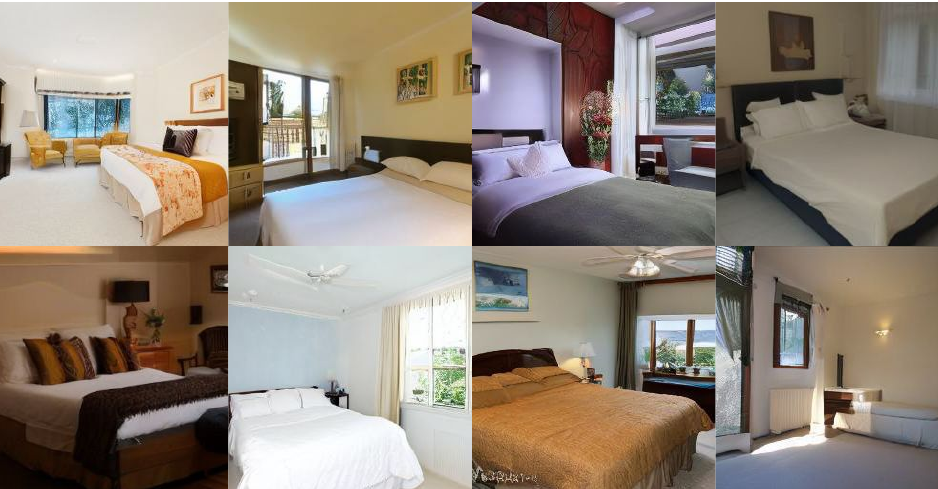}
    \vspace{-0.2in}
        \caption{W4A8}
    \end{subfigure}
    \hfill
    \begin{subfigure}[b]{0.48\textwidth}
    \includegraphics[trim=150 150 120 20,clip,width=\textwidth]{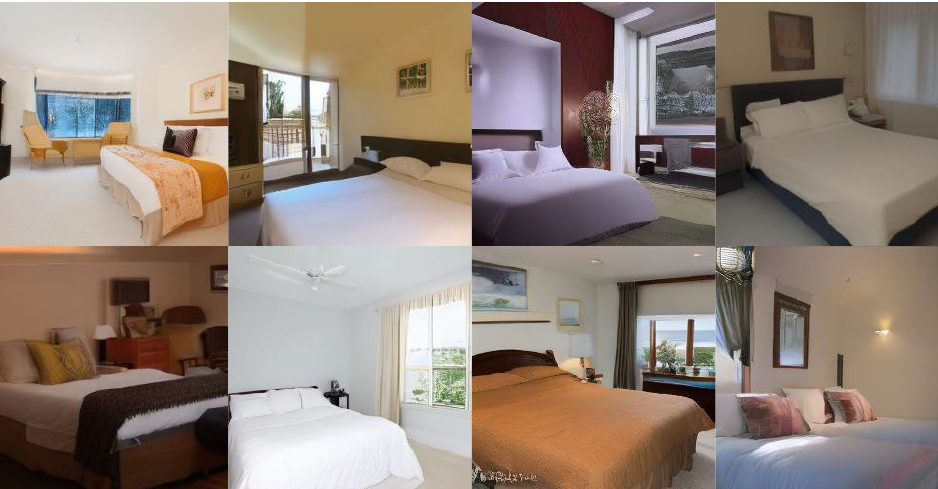}
    \vspace{-0.2in}
        \caption{W4A4}
    \end{subfigure} 
    \vspace{-20pt}
    \caption{Unconditional image generation examples for LSUN-Bedrooms 256$\times$256.}
    \label{fig:bed_samples}
    \vspace{-20pt}
\end{figure}

We see that when corrupting randomly, generation performance is hardly effected. However, corrupting the same amount of tokens (even only one token) with the largest values leads to significantly degenerated images. 

\begin{figure}[h]
    \centering
    \includegraphics[trim=100 0 100 0,clip,width=1.0\linewidth]{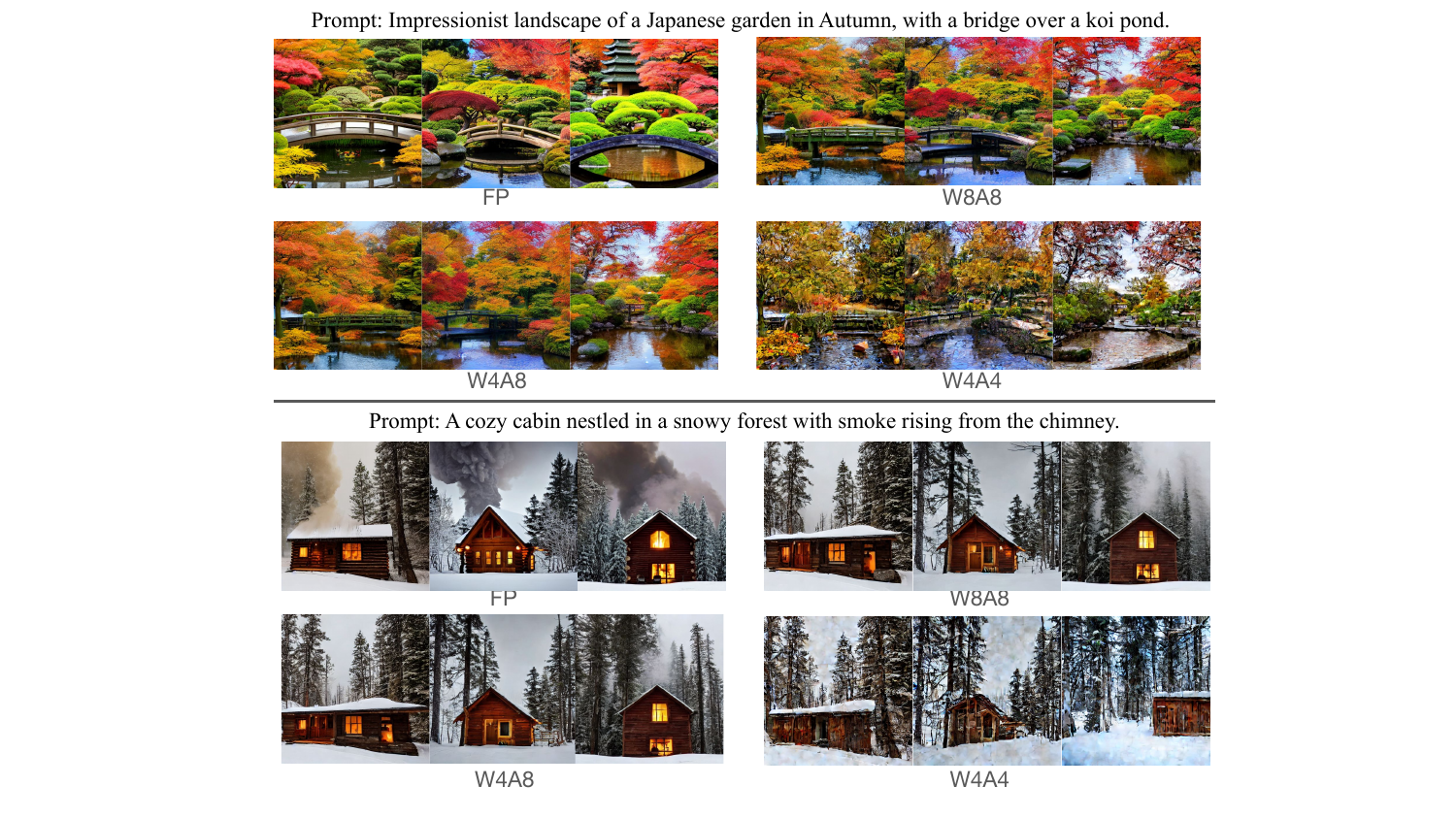}
    \vspace{-28pt}
    \caption{Text-to-image generation results on Stable Diffusion.}
    \label{fig:sd_images}
    \vspace{-10pt}
\end{figure}

\section{More generated image examples}
\label{sec:image_example}
\subsection{Unconditional Image Generation}
\begin{figure}[h]
    \centering
    \begin{subfigure}[b]{0.48\textwidth}
    \includegraphics[trim=150 150 120 20,clip,width=\textwidth]{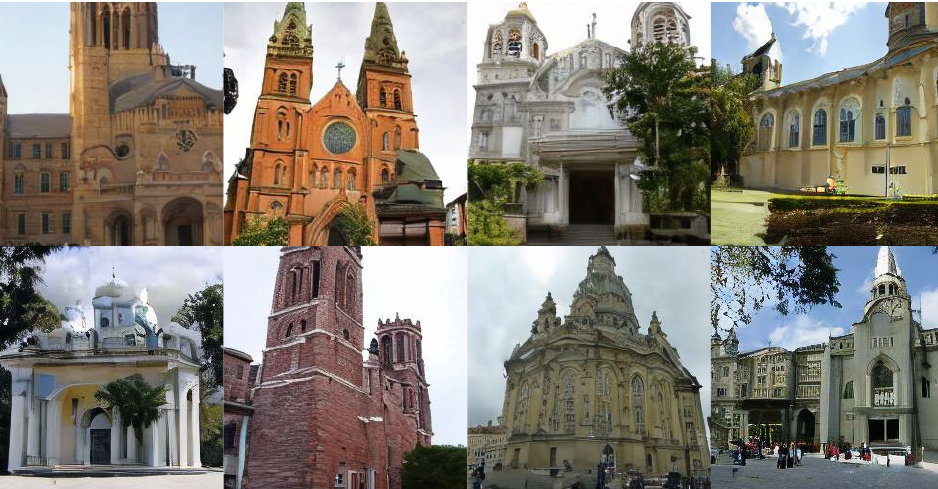}
    \vspace{-0.2in}
        \caption{Full Precision}
    \end{subfigure}
    \hfill
    \begin{subfigure}[b]{0.48\textwidth}
    \includegraphics[trim=150 150 120 20,clip,width=\textwidth]{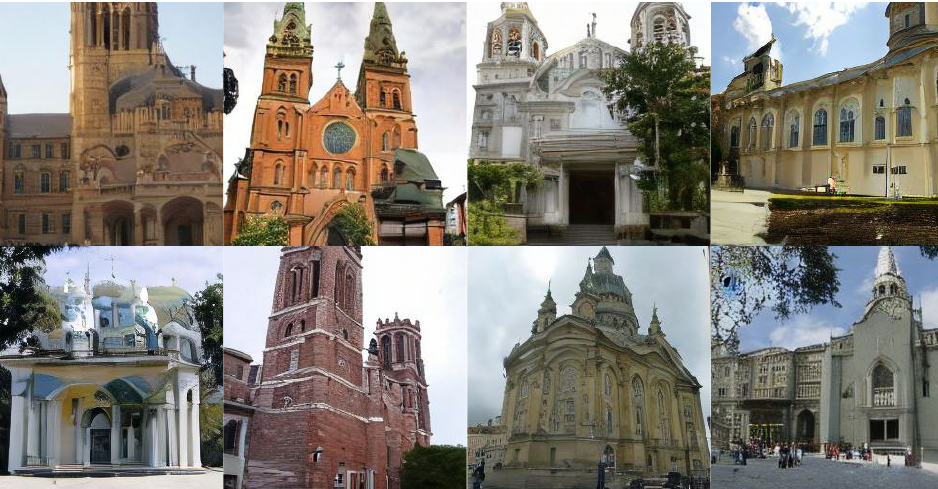}
    \vspace{-0.2in}
        \caption{W8A8}
    \end{subfigure} \\
    \begin{subfigure}[b]{0.48\textwidth}
    \includegraphics[trim=150 150 120 20,clip,width=\textwidth]{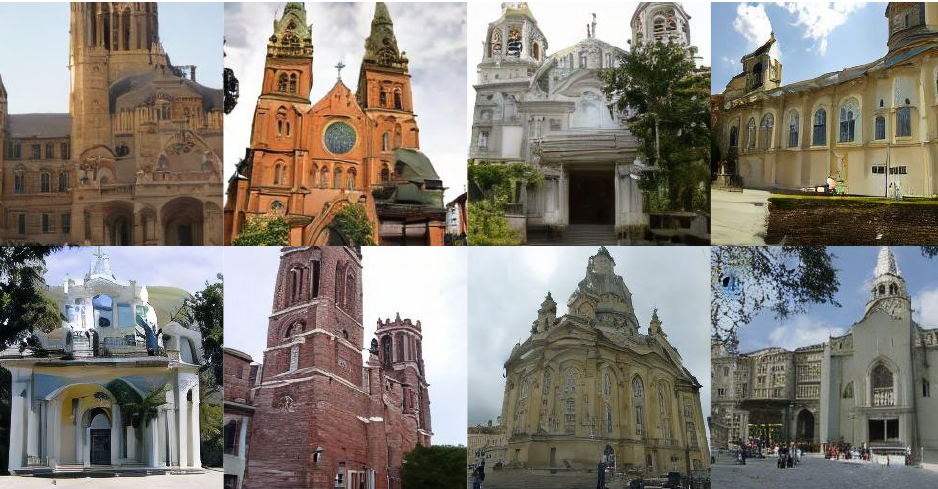}
    \vspace{-0.2in}
        \caption{W4A8}
    \end{subfigure}
    \hfill
    \begin{subfigure}[b]{0.48\textwidth}
    \includegraphics[trim=150 150 120 20,clip,width=\textwidth]{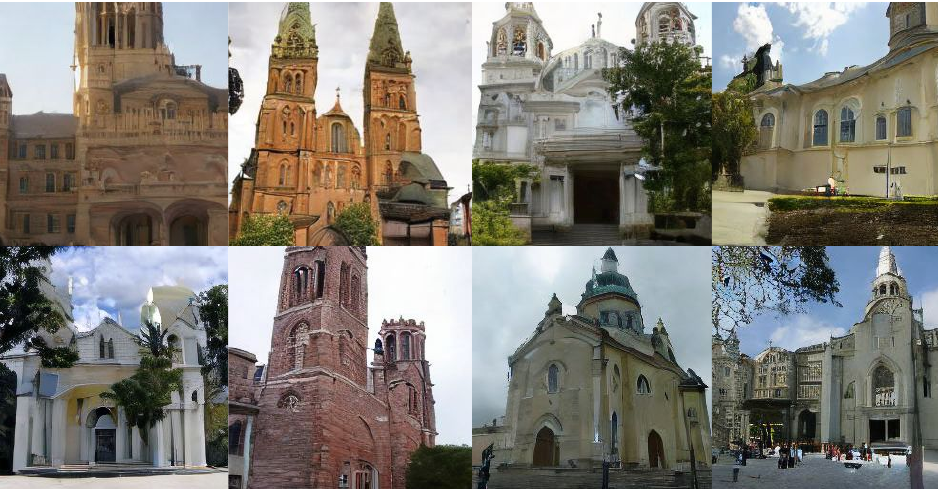}
    \vspace{-0.2in}
        \caption{W4A4}
    \end{subfigure} 
    \vspace{-20pt}
    \caption{Unconditional image generation examples for LSUN-Churches 256$\times$256.}
    \label{fig:church_samples}
    \vspace{-12pt}
\end{figure}

The generated images for LSUN-Bedrooms 256$\times$256 under different bit-widths are shown in \cref{fig:bed_samples}. Images for LSUN-Churches 256$\times$256 are shown in \cref{fig:church_samples}.

\subsection{Class-conditional image generation}
\begin{figure}[h]
    \centering
    \vspace{-4pt}
    \begin{subfigure}[b]{0.48\textwidth}
    \includegraphics[trim=150 150 120 20,clip,width=\textwidth]{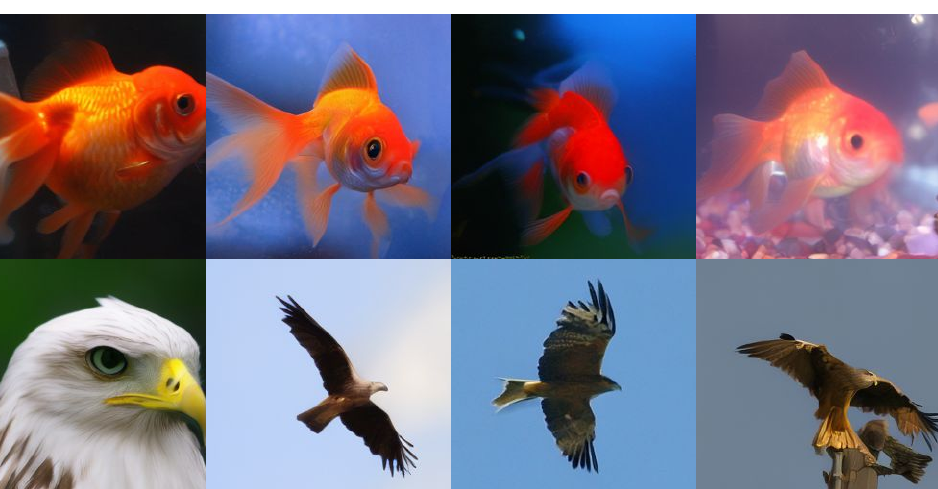}
    \vspace{-0.2in}
        \caption{Full Precision}
    \end{subfigure}
    \hfill
    \begin{subfigure}[b]{0.48\textwidth}
    \includegraphics[trim=150 150 120 20,clip,width=\textwidth]{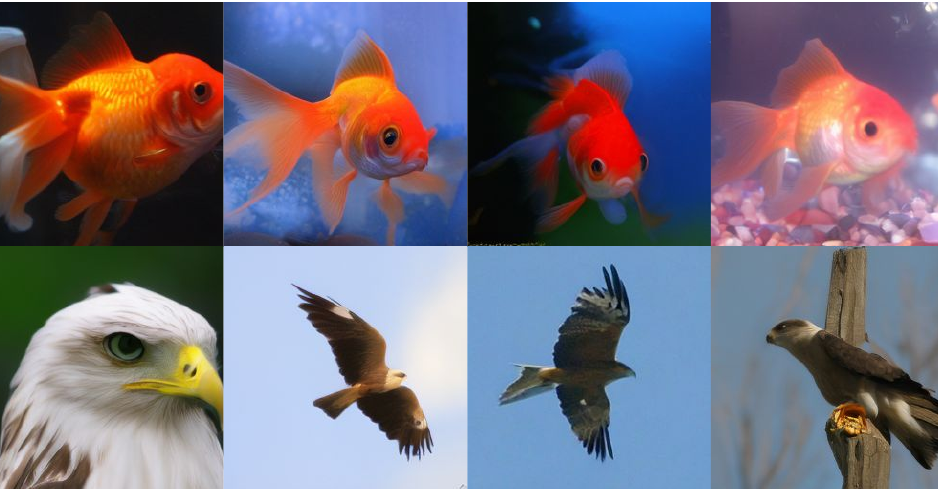}
    \vspace{-0.2in}
        \caption{W8A8}
    \end{subfigure} \\
    \begin{subfigure}[b]{0.48\textwidth}
    \includegraphics[trim=150 150 120 20,clip,width=\textwidth]{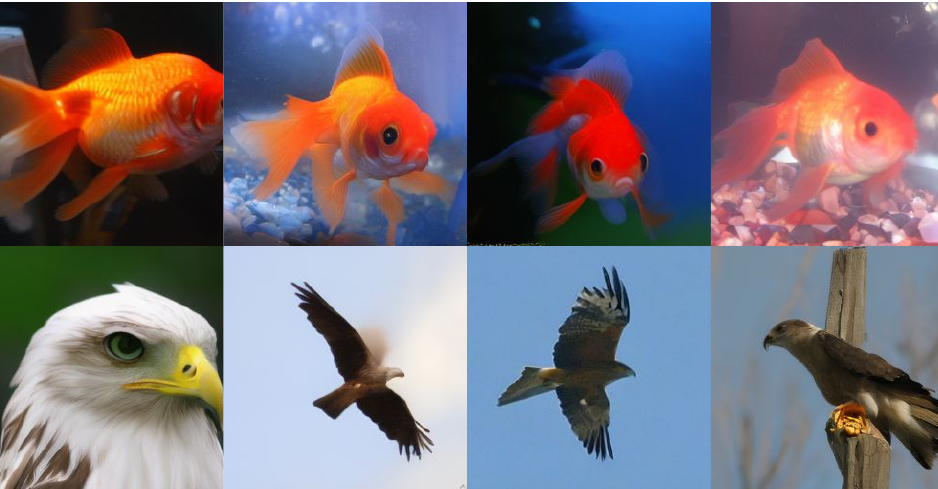}
    \vspace{-0.2in}
        \caption{W4A8}
    \end{subfigure}
    \hfill
    \begin{subfigure}[b]{0.48\textwidth}
    \includegraphics[trim=150 150 120 20,clip,width=\textwidth]{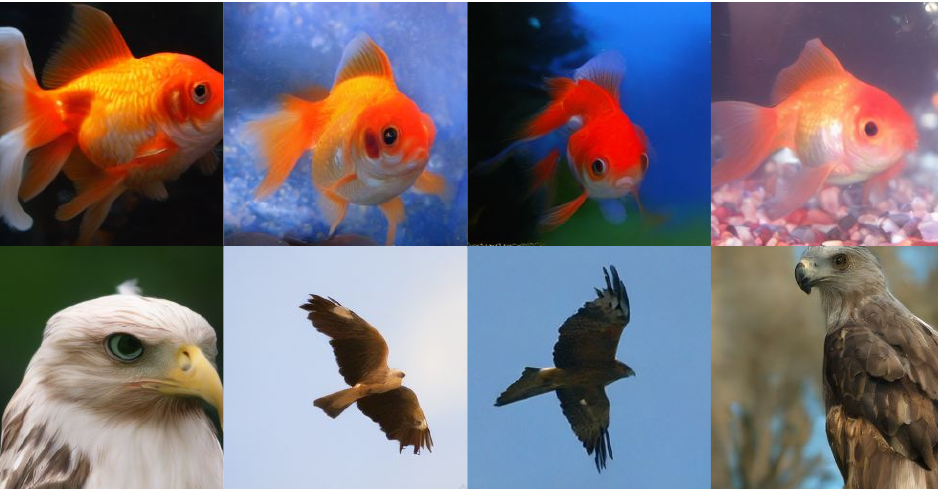}
    \vspace{-0.2in}
        \caption{W4A4}
    \end{subfigure} 
    \vspace{-20pt}
    \caption{Conditional image generation results for ImageNet 256$\times$256.}
    \label{fig:imagenet_samples}
    \vspace{-12pt}
\end{figure}

\cref{fig:imagenet_samples} shows the generated images for 3 different classes. 

\subsection{Text-to-image generation}
\cref{fig:sd_images} shows the generated images using Stable Diffusion v1.4 under different bit-width.

\section{Limitations and Broader Impacts}
\label{sec:limitations}
The primary objective of this paper is to further the research in enhancing the efficiency of diffusion models. While it confronts societal consequences akin to those faced by research on generative models, it is important to recognize the potential impacts that quantized models could have on current techniques, including watermarking and safety checking. Inappropriate integration of current methodologies may result in unforeseen performance issues, a factor that deserves attention and awareness.

\end{document}